\documentclass[lettersize,journal]{IEEEtran}
\usepackage{times}
\usepackage{soul}
\usepackage{url}
\usepackage[hidelinks]{hyperref}
\usepackage[utf8]{inputenc}
\usepackage[small]{caption}
\usepackage{graphicx}
\usepackage{amsmath}
\usepackage{amsthm}
\usepackage{booktabs}
\usepackage{algorithm}
\usepackage{algorithmic}
\usepackage{adforn}
\usepackage{bm}
\usepackage{microtype}
\usepackage{colortbl}
\definecolor{Ocean}{RGB}{129,194,234}

\definecolor{coralred}{rgb}{1.0, 0.25, 0.25}

\definecolor{britishracinggreen}{rgb}{0.12, 0.3, 0.17}

\usepackage{pifont}
\usepackage{hyperref}
\usepackage{multirow}
\usepackage[dvipsnames]{xcolor}
\hypersetup{
    colorlinks=true,
    citecolor = Violet,
    linkcolor = Maroon,
    urlcolor=MidnightBlue
}
\usepackage[switch]{lineno}
\usepackage{amsmath,amsfonts}
\usepackage{algorithmic}
\usepackage{algorithm}
\usepackage{array}
\usepackage[caption=false,font=normalsize,labelfont=sf,textfont=sf]{subfig}
\usepackage{textcomp}
\usepackage{stfloats}
\usepackage{url}
\usepackage{verbatim}
\usepackage{graphicx}
\usepackage{cite}
\usepackage{array} 
\renewcommand\arraystretch{1.2}
\usepackage{enumitem}
\setlist[itemize]{leftmargin=5mm}
\usepackage[switch]{lineno}

\begin{document}

\title{A Survey on Deep Active Learning: Recent Advances and New Frontiers}

\author{Dongyuan Li, 
Zhen Wang,
Yankai Chen,
Renhe Jiang,
Weiping Ding,~\IEEEmembership{Senior Member,~IEEE},
Manabu Okumura
\thanks{Dongyuan Li, Zhen Wang, Manabu Okumura are with the Institute of Innovative Research, School of Information and Communication Engineering, Tokyo Institute of Technology, Tokyo 152-8550, Japan.
(\{lidy,wzh\}@lr.pi.titech.ac.jp, oku@pi.titech.ac.jp). (D. Li and Z. Wang contributed equally to this work).

Yankai Chen is with the School of Computer Science and Engineering, The Chinese University of Hong Kong, (email: ykchen@cse.cuhk.edu.hk).

Renhe Jiang is with the Center for Spatial Information Science, The University of Tokyo. Tokyo, Japan. (email: jiangrh@csis.u-tokyo.ac.jp).

Weiping Ding is with the School of Information Science and Technology, Nantong University, Nantong 226019, China (e-mail: dwp9988@163.com)
}
\thanks{Corresponding authors: Weiping Ding and Manabu Okumura.}}

\markboth{IEEE Transactions on Neural Networks and Learning Systems
}%
{Shell \MakeLowercase{\textit{et al.}}: A Sample Article Using IEEEtran.cls for IEEE Journals}



\maketitle
\begin{abstract}

Active learning seeks to achieve strong performance with fewer training samples.
It does this by iteratively asking an oracle to label new selected samples in a human-in-the-loop manner. 
This technique has gained increasing popularity due to its broad applicability, yet its survey papers, 
especially for deep learning-based active learning (DAL), remain scarce.
Therefore, we conduct an advanced and comprehensive survey on DAL. 
We first introduce reviewed paper collection and filtering.
Second, we formally define the DAL task and summarize the most influential baselines and widely used datasets.
Third, we systematically provide a taxonomy of DAL methods from five perspectives, including annotation types, query strategies, deep model architectures, learning paradigms, and training processes, and objectively analyze their strengths and weaknesses. 
Then, we comprehensively summarize main applications of DAL in Natural Language Processing (NLP), Computer Vision (CV), and Data Mining (DM), etc.
Finally, we discuss challenges and perspectives after
a detailed analysis of current studies.
This work aims to serve as a useful and quick guide for researchers in overcoming difficulties in DAL.
We hope that this survey will spur further progress in this burgeoning field.

\end{abstract}


\section{Introduction}
\begin{figure}[t]
\includegraphics[width=0.48\textwidth]{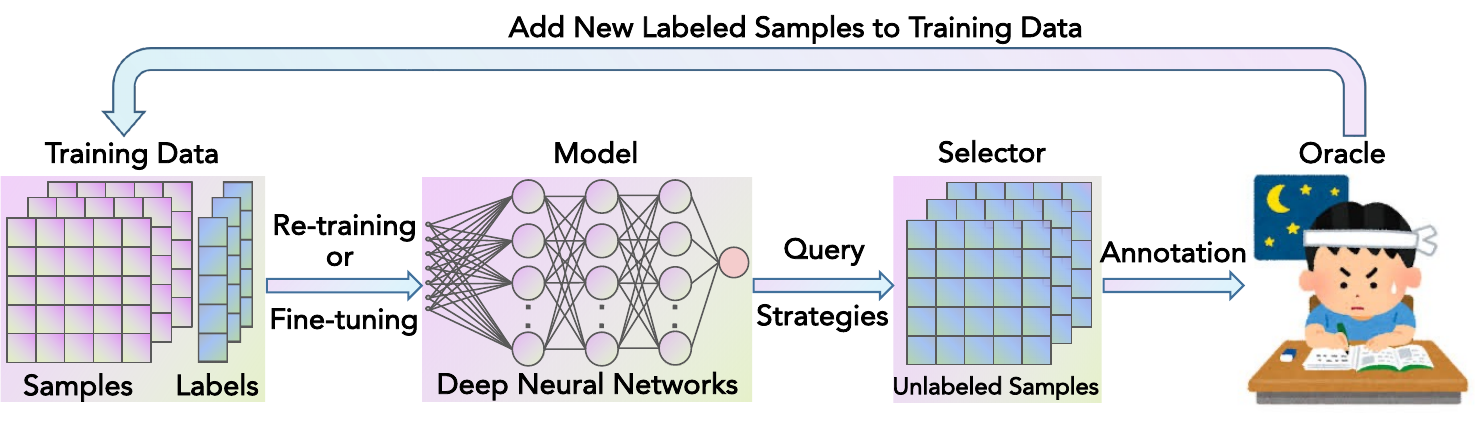}
\caption{The general pipeline in deep active learning.}
\label{pipeline}
\end{figure}

\IEEEPARstart{T}{he} remarkable success of deep learning relies heavily on large-scale datasets with human-annotated labels~\cite{9046288}.
However, continually labeling large-scale datasets is an extremely time-consuming, expensive, and laborious task, which tends to become a bottleneck for deep learning with limited labeled data. 
To tackle this issue, Deep Active Learning (DAL) recently exhibits great potential.
As Fig.~\ref{pipeline} shows, DAL models are first trained on an initial training dataset.
Then, query strategies can be iteratively applied to select the most informative and representative samples from a large pool of unlabeled data. 
Finally, an oracle labels the selected samples and adds them to the training dataset for retraining or fine-tuning of the DAL models. 
DAL aims to achieve competitive performance while reducing annotation costs within a reasonable time~\cite{DBLP:journals/tnn/GuZDH21, DBLP:journals/tnn/CaoT22,DBLP:journals/tnn/LiuXWZYLC23}.
Benefiting from the strong representation capabilities of various neural networks, such as Graph Neural Networks (GNNs)~\cite{DBLP:conf/iclr/KipfW17}, Convolutional Neural Networks (CNNs)~\cite{DBLP:journals/pieee/LeCunBBH98}, and Transformers~\cite{Transformers}, as well as leveraging prior knowledge from pre-trained models like Contrastive Language-Image Pre-Training (CLIP)~\cite{DBLP:conf/icml/RadfordKHRGASAM21} and Generative Pre-trained Transformer (GPT)~\cite{DBLP:journals/corr/abs-2303-08774},
DAL has made significant advances.

As a methodology for selecting or generating a subset of training data in data-centric AI,
DAL is closely related to learning settings and practical techniques, including curriculum learning~\cite{DBLP:conf/icml/BengioLCW09}, transfer learning~\cite{DBLP:journals/tnn/ShaoZL15},
data augmentation or pruning~\cite{DBLP:conf/eccv/ChenZWCL22,DBLP:conf/iclr/0006XP0S023},
and dataset distillation~\cite{DBLP:conf/iccv/PengWLY21}.
The commonality of these methods is to train or fine-tune a model using a small number of samples, aiming to remove noise and redundancy while improving training efficiency without decreasing models' performance on downstream tasks.
However, one primary difference from DAL is that these approaches have {\textit{full access}} to all labels when selecting, distilling, or generating training subsets. 
DAL defaults to that all data should be unlabeled during the training subset selection process, making it better suited for real-world scenarios where labels are initially unavailable.

To summarize DAL methodologies, recent efforts have focused on specific tasks such as text classification~\cite{ DBLP:conf/emnlp/ZhangSH22a} and image analysis~\cite{DBLP:journals/corr/abs-2104-07784,DBLP:journals/corr/abs-2310-14230}, specific domains like NLP~\cite{hadian-sameti-2014-active} and CV~\cite{IA1,IA2}, or reproducing mainstream baselines~\cite{DBLP:conf/ijcai/ZhanL0C21,DBLP:journals/corr/abs-2203-13450}.
As for most early survey work, one common inadequacy is that they may not have enough discussion of recent advances~\cite{DBLP:journals/kais/FuZL13,DBLP:books/crc/aggarwal14/AggarwalKGHY14,math11040820},
or lack summarization of emerging learning paradigms (contrastive learning etc.) and challenges~\cite{B2,B3}, especially in light of rapidly developing deep learning techniques (e.g., Fine-tune on pre-trained models).
To assist researchers in reviewing, summarizing, and planning for future exploration, we provide a comprehensive review encompassing the latest advancements and insights in the field. 
While some survey papers focus on stream-based DAL~\cite{DBLP:journals/ml/CacciarelliK24}, this paper concentrates on pool-based DAL.

\begin{figure}[t]
\centering
\includegraphics[width=0.475\textwidth]{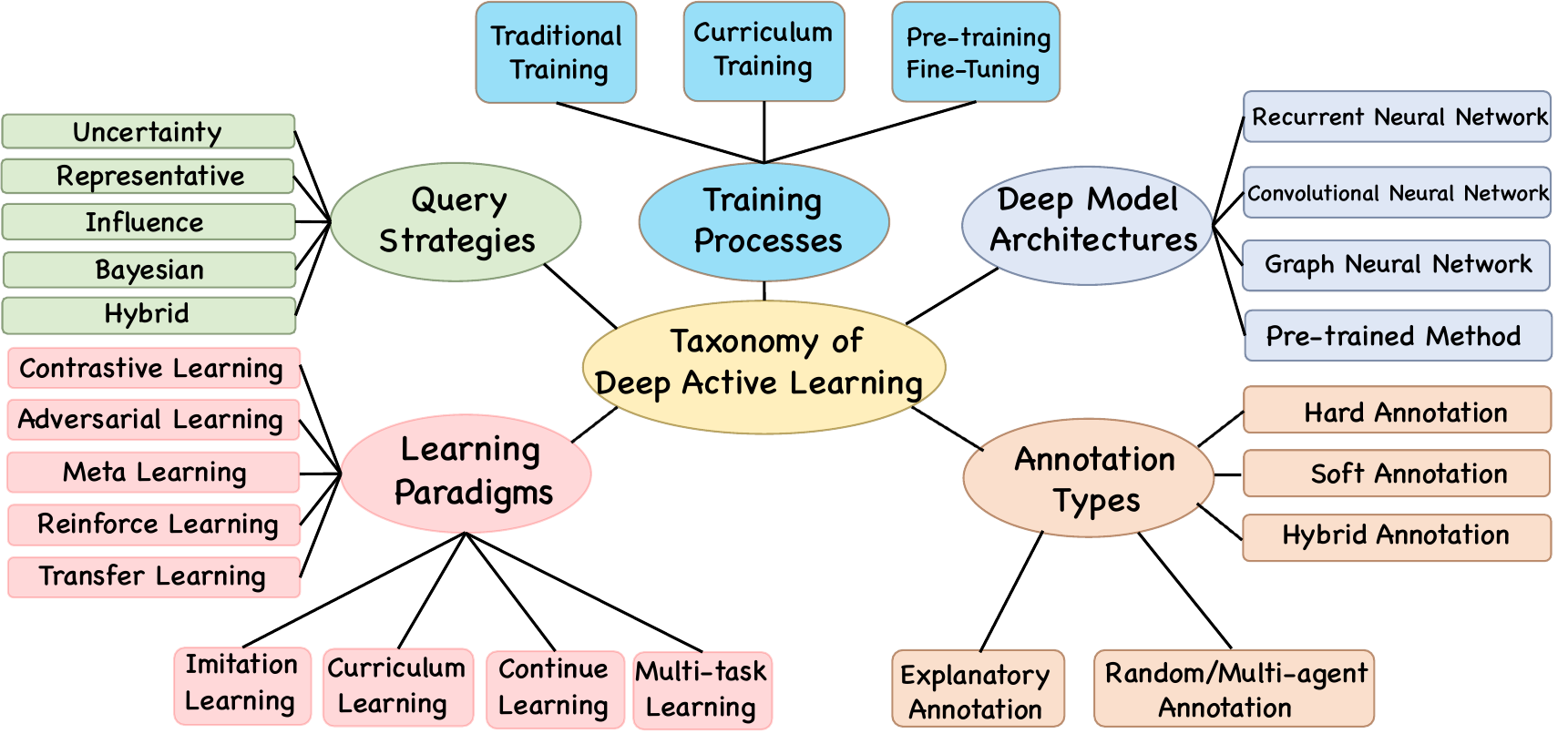}
\caption{Taxonomy for deep active learning methods.}
\label{taxonomy}
\end{figure}

Specifically, we first introduce our strategy for collecting reviewed papers and explain our criteria for selecting them in Section~\ref{sec:collect}. 
Then, we give a specific formal definition for DAL in Section~\ref{notation}, and chronologically summarize the most influential DAL baselines and the widely used datasets in Section~\ref{Import}.
As Fig.~\ref{taxonomy} shows, in Section~\ref{Taxo}, we develop a high-level taxonomy to provide a broad overview of this field, categorizing previous studies from five perspectives.
In Section~\ref{Annotation_Type}, we classify the annotation types into hard, soft, hybrid, explanatory, and random/multi-agent  annotations, and give a detailed introduction to each annotation type.
In Section~\ref{Query_Strategy}, we summarize query strategies into five distinct categories, including uncertainty-based, representative-based, influence-based, Bayesian-based and their hybrid methods, and analyze the strengths and weaknesses of each query type. 
As for deep model architectures, in Section~\ref{deep_arch},  they are mainly categorized into Recurrent Neural Networks (RNNs), CNNs, GNNs, and Pre-trained methods. We discuss the benefits and drawbacks of each type of architecture.  
In Section~\ref{Learning_Paradigm}, we are pleased to discover that various emerging learning paradigms, such as Curriculum Learning and Continual Learning,  have shown promising results when combined with DAL. 
For each learning paradigm, we provide a detailed description of its definition and how to integrate it with DAL. 
Finally, in Section~\ref{Training_Process}, three different training processes, including traditional training, curriculum learning-based training, and pre-training \& fine-tuning will be introduced with typical examples.

\begin{figure}[h]
\centering
\includegraphics[width=0.47\textwidth]{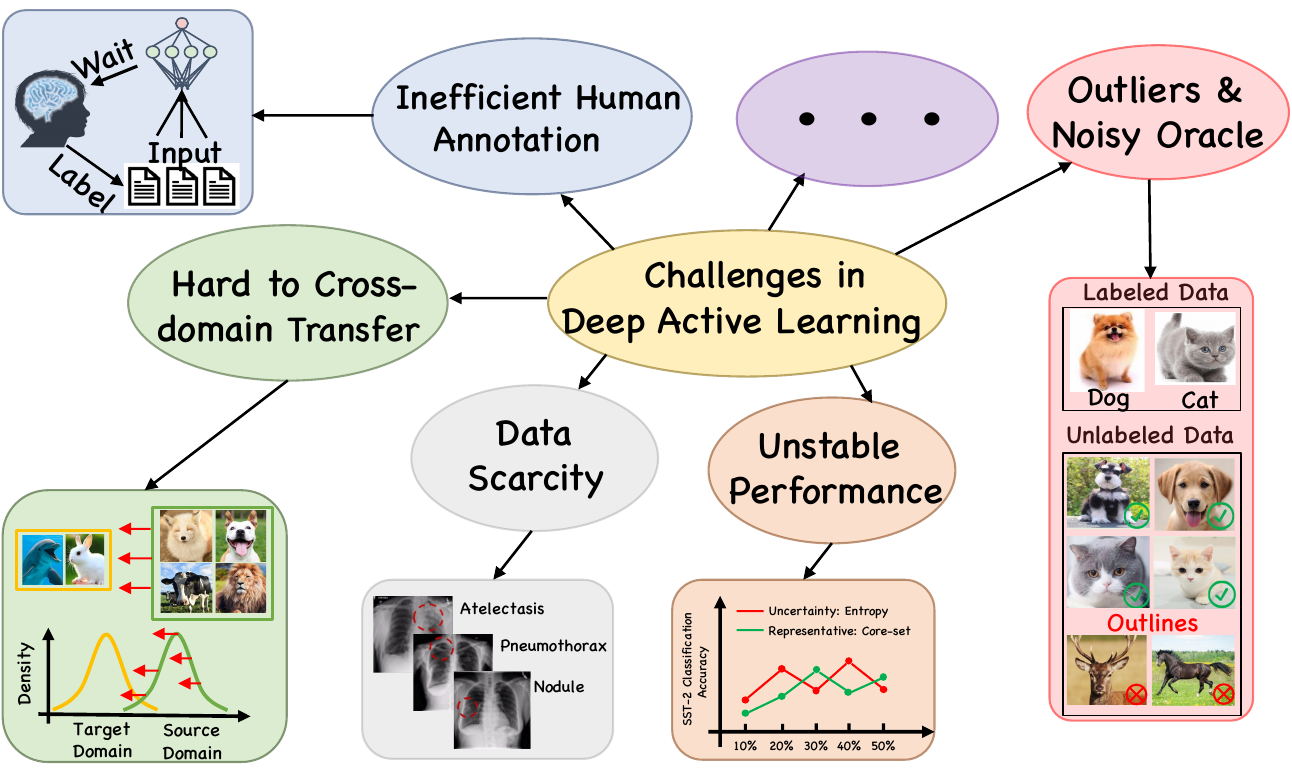}
\caption{Emerging challenges in deep active learning.}
\label{chal}
\end{figure}

In Section~\ref{Application}, we comprehensively show some domains in which DAL methods have been successfully applied, including NLP, CV, DM, etc.
As depicted in Fig.~\ref{chal}, despite the remarkable progress in DAL, this rapidly developing field is still fraught with several crucial emerging challenges.
In Section~\ref{Challenge}, we analyze the causes and opportunities of each challenge, which can be summarized as follows: 
\begin{itemize}
    \item \textbf{Pipeline-related}: inefficient \& costly human annotation, insufficient research on stopping strategies, and cold-start;
    \item  \textbf{Task-related}: difficulty in cross-domain transfer, unstable performance, and lack of scalability and generalizability; 
    \item \textbf{Dataset-related}: outlier data \& oracles, data scarcity \& imbalance, and class distribution mismatch.
\end{itemize}

Finally, after organizing and summarizing the current DAL-related research, we have four intriguing findings that we would like to share with the readers: 
(1) As shown in Section~\ref{Training_Process}, DAL has great potential as a sample selection strategy to apply few-shot or one-shot setting for large-scale pre-trained models with billions of parameters~\cite{margatina-etal-2022-importance,ayub2022fewshot}.
Furthermore, as discussed in Section~\ref{Import}, many studies have shown that using only 10$\sim$20\% labeled samples for fine-tuning the pre-trained language models with billions of parameters can yield even better performance and be 5$\sim$10 times more efficient than training with a full labeled dataset~\cite{DBLP:conf/emnlp/YuanLB20,DBLP:conf/aaai/SeoKAL22}. 
(2) Intuitively, having more high-quality samples can promote model performance for some tasks.
Thus, as shown in Section~\ref{Learning_Paradigm}, many works integrate DAL with semi-supervised strategies, allowing to obtain more high-quality labeled samples without increasing the need for human labor.
However, as discussed in Section~\ref{data-related}, 
semi-supervised methods are highly sensitive to outliers and error labels, easily fueling a vicious cycle, i.e., models continue to label samples with wrong pseudo-labels. 
How to effectively integrate DAL with semi-supervised strategies, using human-labeled true signals to guide semi-supervised annotation and avoid the mislabel circular, remains an open and challenging issue waiting to be solved.
(3) From the detailed analysis of Scalability \& Generalizability in Section~\ref{task-related},
although DAL has achieved great success in classification tasks, comparing various DAL methods to choose the optimal one for a given task remains time-intensive and unrealistic in practice.
Thus, there is an urgent need for a universal framework that is friendly to various downstream tasks. 
(4) By summarizing DAL applications for NLP in Section~\ref{summarization}, we find only a few DAL studies focused on \textit{generative tasks}. 
Generative tasks, such as summarization and question answering, urgently require more attention and research compared to classification tasks. 
This is because generating informative objects, such as annotations, is more difficult and time-consuming. 
Defining the most meaningful samples for generation tasks and explaining why those samples play an important role are two core problems that need to be solved. 
We hope that future research can promote the development of DAL for generation tasks.

Overall, the main contributions of this paper are as follows:
\setlist[itemize]{leftmargin=4mm}
\begin{itemize}
    \item This is the latest comprehensive and systematic survey paper on DAL to help researchers review, summarize, and look forward to the future about DAL.
    \item Based on the novel DAL texonomy, we detail the explanations and discussions of the methodology, ranging from annotation types, query strategies, deep model architectures, learning paradigms,  and training processes. 
    \item The difficult challenges in DAL are presented from multiple perspectives.
    By a detailed analysis of challenges and current studies, we discuss possible advanced solutions for them.
    \item A GitHub repository\footnote{\url{https://github.com/Clearloveyuan/Awesome-Active-Learning}} is available with the most up-to-date DAL techniques, including papers, code, and datasets.
\end{itemize}

%
%


Remaining part of this survey is organized as follows.
Section~\ref{sec:collect} shows the collection of DAL papers.
Section~\ref{DAL:section} introduces important DAL baselines and datasets.
Section~\ref{Taxo} details the taxonomy of DAL methods.
Section~\ref{Application} reviews DAL-related applications.
Section~\ref{Challenge} introduces DAL challenges and opportunities.
Section~\ref{section:conclusion} ends this article with the conclusions.

\section{Paper Collection and Filtering}
\label{sec:collect}

\begin{figure}[h]
\centering
\includegraphics[width=0.5\textwidth]{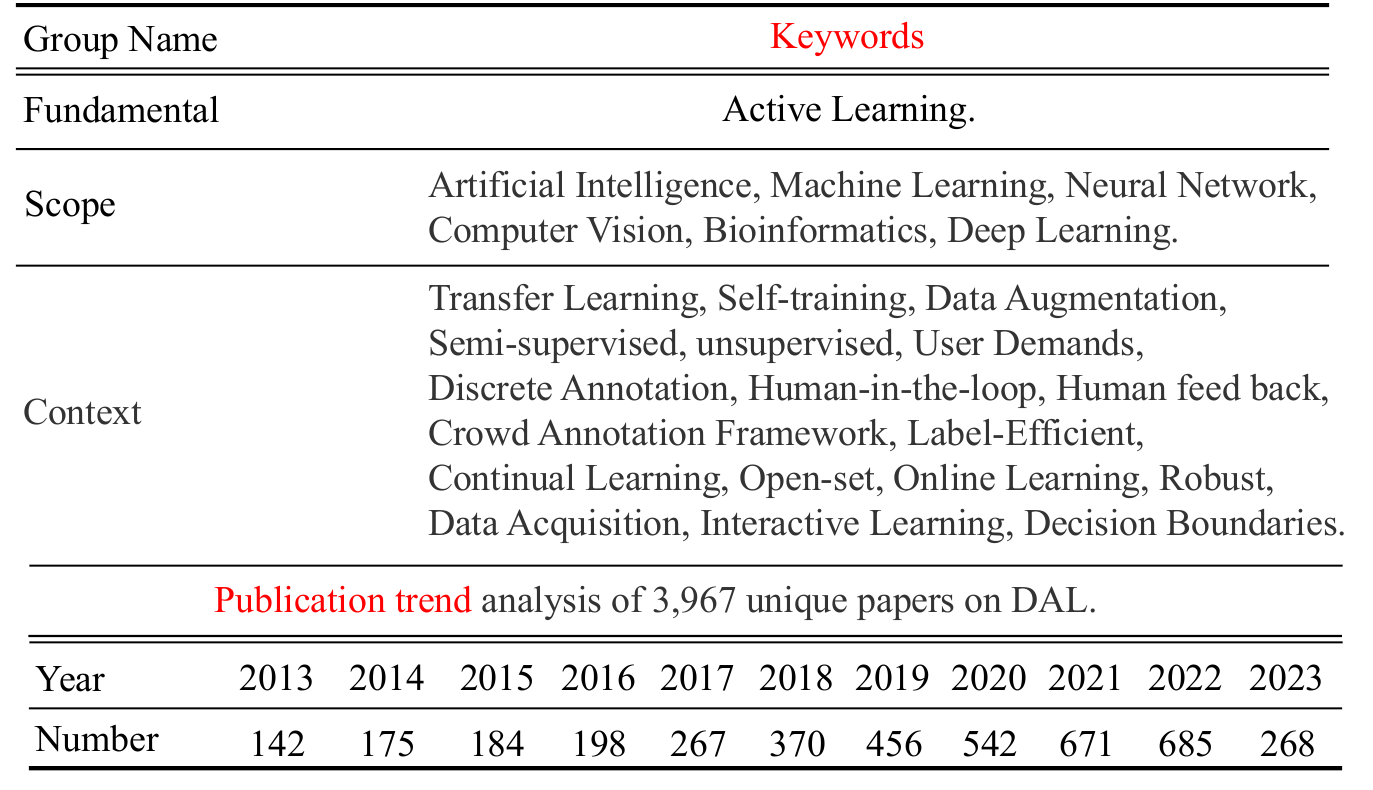}
\caption{Keywords and publication trend on DAL.}
\label{fig:paper_dist}

\end{figure}

We first determine relevant keywords used to search articles and create an initial keyword list, as shown in Fig.~\ref{fig:paper_dist}.
We perform searches across multiple databases using all possible 3-keyword combinations from defined keyword groups, such as ``Active Learning'', ``Machine Learning'', and ``Open-set''. 
The databases searched include Google Scholar, Scopus, Semantic Scholar, and Web of Science. We limit the number of papers collected per query to 200, and the publication date ranges from January 2013 to March 2023.

We collect a total of 10,000 research papers from various sources and obtain 3,967 unique papers after removing any duplicates.
Fig.~\ref{fig:paper_dist} shows the trend of these articles over time, revealing a growing interest in the topic we are investigating.
To ensure the relevance of the collected articles to DAL, we conduct a detailed manual inspection of their abstracts. As a result, we identify 1,273 articles that are considered interesting and pertinent for our study. 
Based on the collected materials, we employ these keywords to perform a final filtering process and also consider the reputation of conferences or journals in which the papers were published, as well as their impact. 
This approach further refines our dataset, resulting in 405 articles that are selected for systematic analysis, and 220 articles are finally summarized and discussed, focusing on their key findings and contributions. 
This rigorous analysis ensures that the articles are relevant and provide valuable insight into the field of DAL.

\section{Deep Active Learning}
\label{DAL:section}

In this section, we first introduce the basic notation and definition of DAL and then discuss the most important DAL baselines based on their relevance and chronological order.

\begin{algorithm}[h]
\small
    \caption{DAL procedure.}
    \label{deep_algorithm}
    \textbf{Input}: Unlabeled Data $\mathcal{D}_{\textbf{pool}}$\\
    \textbf{Parameter}: Batch Size $b$, Iteration Times $T$, Query Function $\alpha$ \\
    \textbf{Output}: The final trained model $\mathcal{M}$
    \begin{algorithmic}[1] 
        \STATE $\mathcal{Q}_{0} \leftarrow$ Initialization sampling from $\mathcal{D}_{\textbf{pool}}$ where $|\mathcal{Q}_{0}|=b$;
        \STATE $\mathcal{D}_{\textbf{train}}^{0} \leftarrow \mathcal{Q}_{0}$ [Initialization of training dataset];
        \STATE $\mathcal{M}_{0} \leftarrow$ Train $\mathcal{M}_{0}$ on $\mathcal{D}_{\textbf{train}}^{0}$; 
        \WHILE{\textbf{not} stop-criterion(\,) $\&$ i $\leq$ $T$}
        \STATE $\mathcal{Q}_{i}$ $\leftarrow$ $\alpha(\mathcal{M}_{i-1}, \mathcal{D}_{\textbf{pool}}^{i-1},b$) \,\,[Annotating $b$ samples];
        \STATE $\mathcal{D}_{\textbf{train}}^{i}$\,\,=\,\,$\mathcal{D}_{\textbf{train}}^{i-1}$ $\cup$ $\mathcal{Q}_{i}$; \,\,\,\, $\mathcal{D}_{\textbf{pool}}^{i} \leftarrow \mathcal{D}_{\textbf{pool}}^{i-1} \backslash \mathcal{Q}_{i}$;
        \STATE $\mathcal{M}_{i} \leftarrow$ Train  $\mathcal{M}_{i-1}$ on $\mathcal{D}_{\textbf{train}}^{i}$; 
        \ENDWHILE
    \end{algorithmic}
\end{algorithm}

\subsection{\textbf{Notations \& Definition}}
\label{notation}

We focus on pool-based DAL methods since most DAL methods belong to this category. 
Pool-based DAL methods iteratively select the most informative samples from a large pool of unlabeled datasets until either the base model reaches a certain level of performance or a pre-defined budget is exhausted.
As shown in Algorithm~\ref{deep_algorithm}, we use a classification task as an example for illustration, while other tasks follow the typical definition of their task domains. 
Given an initial labeled training dataset $\mathcal{D}_{\textbf{train}}=\{\bm{x}_{i},y_{i}\}_{i=1}^{m}$ and a large-scale pool of unlabeled data $\mathcal{D}_{\textbf{pool}}=\{\bm{x}_{i}\}_{i=1}^{n}$,
where m$\ll$n, $\bm{x}_{i}$ represents the feature vector of the $i$-th sample, and $y_{i} \in \{0,1\}$ is the class label for binary classification (or $y_{i} \in \{1,\dots,k\}$ for multi-label classification),
the {DAL} procedure is carried out in $T$ iterations. 
In the $i$-th iteration, a batch of samples $\mathcal{Q}^{i}$ with batch size $b$ is selected from $\mathcal{D}_{\textbf{pool}}^{i-1}$ on the basis of the base model $\mathcal{M}$ and an acquisition function $\alpha(\,)$.
These samples $\mathcal{Q}^{i}$ are then labeled by an oracle and added to the $i$-th training dataset 
$\mathcal{D}^{i}_{\textbf{train}}$, with which the model $\mathcal{M}$ is then re-trained. 
DAL terminates when the labeled budget $Q$ is exhausted or the desired performance of the model is reached.

\subsection{\textbf{Comparisons between Traditional and Deep AL}}
The differences between traditional and Deep AL mainly lie in the following two aspects: 
(1) most traditional AL methods use fixed pre-processed features to calculate uncertainty/representativeness. In deep learning tasks, feature representations are jointly learned with Deep Neural Networks (DNNs). Therefore, feature representations dynamically change during DAL processes, and thus pairwise distances/similarities used by representativeness-based measures need to be re-computed in every stage.
In contrast, for traditional AL with classical ML tasks, these pairwise terms should be pre-computed~\cite{DBLP:journals/corr/abs-2203-13450}.
(2) DAL can leverage advanced large-scale pre-trained language models to achieve comparable performance in few-shot or one-shot settings. In contrast, traditional AL methods with few-shot or one-shot settings may not meet the minimum requirements for the number of training samples needed to achieve comparable performance~\cite{ayub2022fewshot,DBLP:conf/emnlp/MargatinaSAD23}.
On the other hand, the most similar aspect between traditional and deep AL methods is their utilization of a small number of the most informative samples to train models, thereby improving efficiency and reducing reliance on labeled samples.

\subsection{\textbf{Important DAL Baselines and Datasets}}
\label{Import}

The most important baselines for DAL are carefully categorized in Table~\ref{category} from six perspectives to provide readers with a complete understanding of the development of DAL and the identification of the most relevant works.
These influential studies have achieved breakthroughs in designing new DAL methods, tackling novel tasks, or integrating with emerging learning paradigms.
They have been published in influential international conferences or high-quality journals in machine learning, CV, NLP, etc., and have been highly cited with more than 100 total citations or more than 10 citations per year.

\begin{table*}[t]
\caption{Detailed taxonomy of important Deep Active Learning baselines. Refer to Section~\ref{Taxo} for a detailed explanation of each category. \textbf{Any Types} in Query Strategy means the proposed frameworks can be combined with any types of DAL query strategies.} 
\centering
\footnotesize
\setlength{\tabcolsep}{1.6mm}{
\begin{tabular}{lcccccc}
\toprule
Method & Query Strategy  & Architecture & Learning Paradigm & Annotation & Training & Tasks  \\ \midrule
 \textbf{BCBA [2016]}~\cite{DBLP:journals/corr/GalG15a}  &Bayesian &CNNs &Traditional & Hard & Traditional &Image Classification \\
 \textbf{DBAL [2017]}~\cite{DBLP:conf/icml/GalIG17} &Bayesian &CNNs &Semi-supervised Learning & Hard  & Traditional & Image Classification \\
 \textbf{CEAL [2017]}~\cite{DBLP:journals/tcsv/WangZLZL17} &Uncertainty &CNNs &Curriculum Learning &Hybrid   & Curriculum &Image Classification  \\
 \textbf{ESNN [2017]}~\cite{DBLP:conf/nips/Lakshminarayanan17} &Uncertainty &BNNs &Adversarial Learning & Hard  & Traditional &Image Classification \\
\textbf{PAL [2017]}~\cite{DBLP:conf/emnlp/FangLC17} &Uncertainty &BNNs & Reinforcement Learning  & Hard  & Traditional & Named Entity Recognition\\
\textbf{LAL [2017]}~\cite{DBLP:conf/nips/KonyushkovaSF17} &Influence &Random Forest  &Traditional  & Hard   & Traditional & Regression Tasks\\
\textbf{GAAL [2017]}~\cite{DBLP:journals/corr/ZhuB17} & Uncertainty & GNNs & Adversarial Learning & Hard   & Traditional & Image Classification \\ 
\textbf{CoreSet [2018]}~\cite{sener2018active} &Representative &CNNs &Semi-supervised Learning & Hard  & Traditional &Image Classification\\
\textbf{DFAL [2018]}~\cite{DBLP:journals/corr/abs-1802-09841}&Uncertainty &CNNs &Adversarial {Training} & Hard  & Traditional &Image Classification\\
 \textbf{ASM [2019]}~\cite{DBLP:journals/tnn/WangLYCZZ19} & Uncertainty & CNNs & Curriculum {Learning}  & Hybrid   & Curriculum & Objective Detection \\
 \textbf{MIAL [2019]}~\cite{DBLP:journals/tnn/CarbonneauGG19} &Representative &  SVM & Traditional & Hard   & Traditional & Image Classification\\
 \textbf{BatchBALD [2019]}~\cite{DBLP:conf/nips/KirschAG19}  & Uncertainty & BNNs & Traditional & Hard   & Traditional & Image Classification\\
 \textbf{DRAL [2019]}~\cite{DBLP:conf/iccv/LiuWGTL19} & Uncertainty &CNNs &Reinforcement Learning &Hard   &Pre+FT & Person Re-Identification  \\
\textbf{DLER [2019]}~\cite{DBLP:conf/acl/KasaiQGLP19} & Uncertainty & PLMs & Transfer Learning  & Hard   & Pre+FT & Entity Resolution\\
\textbf{BGADL [2019]}~\cite{DBLP:conf/icml/TranDRC19} & Hybrid & BNNs & Semi-supervised Learning & Hard  & Traditional & Image Classification \\
\textbf{VAAL [2019]}~\cite{DBLP:conf/iccv/SinhaED19} & Representative & VAE & Adversarial Learning & Hard   & Traditional & Image Classification  \\
\textbf{AADA [2020]}~\cite{DBLP:conf/wacv/SuTSLMC20} & Hybrid & CNNs & Transfer Learning  & Hard   & Pre+FT & Object Detection \\
\textbf{CSAL [2020]}~\cite{DBLP:conf/eccv/GaoZYADP20} & Hybrid & CNNs & Traditional & Hard   & Pre+FT & Image Classification \\
\textbf{SRAAL [2020]}~\cite{DBLP:conf/cvpr/Zhang0YWZH20} & Uncertainty & CNNs & Adversarial Learning & Hard  & Pre+FT & Image Classification \\
\textbf{ALPS [2020]}~\cite{DBLP:conf/emnlp/YuanLB20}& Uncertainty & PLMs & Traditional & Hard  & Pre+FT & Cold-start Issue \\
\textbf{Ein-Dor et al. [2020]}~\cite{DBLP:conf/emnlp/Ein-DorHGSDCDAK20}& Any Types & PLMs & Traditional & Hard   & Pre+FT & Text Classification \\
\textbf{TOD [2021]}~\cite{DBLP:conf/iccv/HuangWXHD21} & Uncertainty & CNNs & Traditional & Hard   & Pre+FT & Image Classification \\
\textbf{Cluster-Margin [2021]}~\cite{DBLP:conf/nips/CitovskyDGKRRK21} & Representative & CNNs & Traditional & Hard   & Pre+FT & Image Classification\\
\textbf{LADA [2021]}~\cite{DBLP:conf/nips/KimSJM21} & Uncertainty &CNNs & Semi-supervised Learning & Hard    &Traditional & Image Classification \\
\textbf{TA-VAAL [2021]}~\cite{DBLP:conf/cvpr/KimPKC21} & Influence & VAE & Adversarial Learning & Hard   & Pre+FT & Image Classification \\
\textbf{Karamcheti et al. [2021]}~\cite{Outliers-ACL1}  & Hybrid & PLMs & Traditional & Hard   & Pre+FT & Visual Question Answering \\
\textbf{MAML [2022]}~\cite{zilongzhu} & Any Types & PLMs & Meta Learning  & Hard   & Pre+FT & Text Classification \\
\textbf{BATL [2022]}~\cite{DBLP:conf/aaai/SeoKAL22}  & Hybrid & PLMs & Traditional & Hard   & Pre+FT & Text Classification \\
\textbf{TYROGUE [2022]}~\cite{Maekawa} & Hybrid & PLMs & Traditional & Hard   & Pre+FT & Text Classification \\
\textbf{Schroder et al. [2022]}~\cite{DBLP:conf/acl/SchroderNP22} & Uncertainty & PLMs & Traditional  & Hard  & Pre+FT & Text Classification \\
\bottomrule
\end{tabular}}
\label{category}

\end{table*}

\textbf{BCBA}~\cite{DBLP:journals/corr/GalG15a} pioneers the combination of AL with Bayesian neural networks (BNNs), using Monte Carlo dropout for a variational Bayesian approximation to apply for image classification.
Based on this,
\textbf{DBAL}~\cite{DBLP:conf/icml/GalIG17} proposes an uncertainty-based query strategy for high-dimensional image classification.
To expand number of labeled samples without increasing human labors, \textbf{CEAL}~\cite{DBLP:journals/tcsv/WangZLZL17} combines DAL with semi-supervised strategies by assigning pseudo-labels to high-confidence samples while requesting annotations for the most uncertain samples.
Relying on a single query strategy may lead to errors. Thus, \textbf{ESNN}~\cite{DBLP:conf/nips/Lakshminarayanan17} uses a deep ensemble of DNNs to measure sample uncertainty from multiple aspects and achieves good robustness for unbalanced datasets. 
However, the aforementioned methods are criticized for being less effective for batch DAL~\cite{DBLP:conf/nips/KirschAG19}. 
To address this issue, \textbf{CoreSet}~\cite{sener2018active} selects informative batches that cover the whole data distribution and \textbf{BatchBALD}~\cite{DBLP:conf/nips/KirschAG19} uses mutual information to identify the most informative batches.
And \textbf{Cluster-Margin}~\cite{DBLP:conf/nips/CitovskyDGKRRK21} aims to select informative and diverse mini batches to improve accuracy and efficiency.

To better help DAL adjust to different tasks, reinforcement learning provides detailed rewards for dynamically controlling query strategies. For example, \textbf{PAL}~\cite{DBLP:conf/emnlp/FangLC17} learns a deep reinforcement learning-based Q-network as an adaptive policy to select data samples for labeling. 
Similarly,
\textbf{DRAL}~\cite{DBLP:conf/iccv/LiuWGTL19} uses a reinforcement learning framework to dynamically adjust the acquisition function via rewards to obtain high-quality queries. 
\textbf{UCBVI}~\cite{DBLP:conf/icml/MenardDJKLV21} provides a new modification to the Q-network formulation for reward-free exploration, significantly reducing query complexity. 
However, reinforcement learning requires a large amount of training data and human-designed rewards, which is difficult for many real-world applications. 
To address this issue, meta learning and transfer learning have become main solutions.
\textbf{LAL}~\cite{DBLP:conf/nips/KonyushkovaSF17} trains a regressor to learn optimal query strategies for downstream tasks.
\textbf{MAML}~\cite{zilongzhu} combines meta learning and DAL by initializing an active learner with meta-learned parameters obtained through meta-training on tasks similar to the target task during DAL. 
\textbf{DLER}~\cite{DBLP:conf/acl/KasaiQGLP19} designs an architecture to learn a transferable model from a high-resource setting to a low-resource one, allowing DAL to select a few informative samples based on the knowledge of the source domain. 
\textbf{AADA}~\cite{DBLP:conf/wacv/SuTSLMC20} jointly considers domain alignment, uncertainty, and diversity for sample selection.

To enlarge the labeled training dataset for DNNs without incurring additional human labor costs, semi-supervised, semi-supervised, and self-supervised DAL methods have been proposed.
\textbf{MIAL}~\cite{DBLP:journals/tnn/CarbonneauGG19} pioneers semi-supervised DAL using cluster-based strategies to measure sample informativeness.
\textbf{ASM}~\cite{DBLP:journals/tnn/WangLYCZZ19} collaborates with self-learning and DAL, 
designing a selector function to selectively and seamlessly determine the confidence of the samples, where high-confidence samples are labeled by a pseudo-labeling module, and low-confidence samples are labeled by humans.
\textbf{CSAL}~\cite{DBLP:conf/eccv/GaoZYADP20} first uses semi-supervised learning to distill information from unlabeled data during the training stage and then uses consistency-based sample selection for DAL.
\textbf{TOD}~\cite{DBLP:conf/iccv/HuangWXHD21} leverages a novel unlabeled data sampling strategy for data annotation in conjunction with a semi-supervised training scheme to improve the performance of the task model with unlabeled data. 
Recently, data augmentation has expanded to become a deep neural model that generates virtual instances to help expand training datasets. 
\textbf{GAAL}~\cite{DBLP:journals/corr/ZhuB17} introduces a generative adversarial network to the DAL query method to generate informative samples to train the model. 
\textbf{BGADL}~\cite{DBLP:conf/icml/TranDRC19}  expands GAAL and combines generative adversarial DAL with Bayesian data augmentation to generate diverse and informative samples. 
\textbf{DFAL}~\cite{DBLP:journals/corr/abs-1802-09841} uses adversarial DAL to select samples close to the decision boundary as the most informative samples for DAL.
\textbf{VAAL}~\cite{DBLP:conf/iccv/SinhaED19} learns a latent space using a variational autoencoder (VAE) to generate new informative samples and trains an adversarial network to discriminate labeled and unlabeled data.
Inspired by these works,
\textbf{TA-VAAL}~\cite{DBLP:conf/cvpr/KimPKC21} incorporates a learning loss prediction module and a task ranker to enable task-aware sample selection. 
\textbf{SRAAL}~\cite{DBLP:conf/cvpr/Zhang0YWZH20} proposes a relabel adversarial model that aims to obtain the most informative unlabeled samples.
\textbf{LADA}~\cite{DBLP:conf/nips/KimSJM21} anticipates data augmentation impact by scoring both real and virtually augmented instances, allowing training in informative labeled and augmented data.

Large-scale pre-trained language models (PLMs) achieve great success and become a milestone in artificial intelligence.
Due to sophisticated pre-training objectives and huge model parameters, large-scale PLMs effectively captures knowledge from massive labeled and unlabeled data. 
DAL also ushers in a new paradigm by leveraging the prior knowledge in PLMs to enable few-shot or zero-shot learning for many downstream tasks. 
\textbf{ALPS}~\cite{DBLP:conf/emnlp/YuanLB20} extracts knowledge from PLMs to select the first batch of data using masked language modeling loss, which successfully solves the cold-start problem of DAL. \textbf{Ein-Dor et al.}~\cite{DBLP:conf/emnlp/Ein-DorHGSDCDAK20} use multiple DAL methods to select samples for fine-tuning in BERT-based text classification. It achieves comparable or higher performance than fine-tuning on full datasets only with 10\%$\sim$20\% labeled samples.
\textbf{Karamcheti et al.}~\cite{Outliers-ACL1} use DAL to identify and remove noisy data, select balanced samples to fine-tune PLMs, and achieve better performance in visual question-answering. 
\textbf{BATL}~\cite{DBLP:conf/aaai/SeoKAL22} is a task-independent batch acquisition method on a PLMs with triplet loss to determine hard samples, which have similar features but difficult to identify labels in an unlabeled data pool. 
\textbf{TYROGUE}~\cite{Maekawa} designs an interactive DAL framework to flexibly select samples to fine-tune PLMs for multiple low-resource tasks. 
\textbf{Schroder et al.}~\cite{DBLP:conf/acl/SchroderNP22} extend the PLMs using available unlabeled data for greater adaptability and introduce effective fine-tuning for the robustness of DAL in low-resource and high-resource settings.

As shown in Table~\ref{dataset}, we also conclude the most widely used datasets in DAL including images, text, and audio.

\renewcommand\arraystretch{1}
\begin{table}[t]
\caption{Widely used DAL dataset {information.}}
\centering
\begin{tabular}{lccc}
\toprule
\textbf{Dataset} & \textbf{Size} & \textbf{Domain} & \textbf{Tasks} \\
\midrule
\midrule
MNIST~\cite{DBLP:journals/pieee/LeCunBBH98} & 70,000 & Images & Classification \\
CIFAR-10~\cite{cifar10} & 60,000 & Images & Classification \\
SVHN~\cite{svhn} & 600,000 & Images & Classification, Localization \\
ImageNet~\cite{imagenet} & 1.2M & Images & Classification, Detection \\
MSCOCO~\cite{mscoco} & 123,287 & Images & Object detection \\
Cityscapes~\cite{cityscapes} & 5,000 & Images & Semantic segmentation \\
Caltech-101~\cite{caltech101} & 9,000 & Images & Classification \\
SST~\cite{sst} & 11,855 & Text & Sentiment analysis \\
TREC~\cite{trec} & 5,952 & Text & Question answering \\
SNLI~\cite{snli} & 570,000 & Text & Natural language inference \\
IMDB~\cite{imdb} & 50,000 & Text & Sentiment analysis \\
AGNews~\cite{agnews} & 31,900 & Text & Classification \\
PubMed~\cite{pubmed} & 19,717 & Text & Document classification \\
YouTube-8M~\cite{youtube8m} & 237,000 & Audio & Classification \\
MIMIC-III~\cite{mimic3} & 112,000 & Medical & Healthcare analytic \\
\bottomrule
\end{tabular}
\label{dataset}

\end{table}

\section{Taxonomy of DAL}
\label{Taxo}

\subsection{\textbf{Annotation Type}}
\label{Annotation_Type}

\noindent  {\adfhalfrightarrowhead} \textbf{\textit{Hard annotations}} provide one or multiple discrete categorical labels independently for each sample.
For example, Citovsky et al.~\cite{DBLP:conf/nips/CitovskyDGKRRK21} annotate each image with a specific label such as ``balloon'' or ``strawberry'' for an image classification task.
Wiechman et al.~\cite{DBLP:conf/naacl/WiechmannYB21} design an online annotation system to assign multiple labels to long documents based on their sentiments, topics, and spam/non-spam status. 

\noindent  {\adfhalfrightarrowhead} \textbf{\textit{Soft annotations}} allow continuous and subjective labels for samples. 
For instance, ReDAL~\cite{DBLP:conf/iccv/WuLHLSHH21} annotate continuous 2D region labels for 3D point clouds in semantic segmentation. 
Kothawade et al.~\cite{DBLP:conf/eccv/KothawadeGSXI22} use mutual information as an auxiliary metric to select annotation regions in images for autonomous vehicles. 
Xie et al.~\cite{DBLP:conf/cvpr/XieY0LC22} propose a region-based approach to automatically query a small subset of image regions to label while maximizing segmentation performance.

\noindent {\adfhalfrightarrowhead} \textbf{\textit{Hybrid annotations}} combine automatic pseudo-labels of high-confidence predictions with human labeling of low-confidence samples in an iterative self-paced manner~\cite{DBLP:journals/tnn/WangLYCZZ19}. 
For example, Wang et al.~\cite{DBLP:journals/tcsv/WangZLZL17} propose a complementary sample selection strategy to progressively choose the most informative samples, pseudo-labeling high-confidence predictions for training.
Yu et al.~\cite{DBLP:journals/tnn/YuTWDZ21} jointly use the expertise of different annotation groups, inter-relations between workers, and label correlations within groups. By weighting groups, they reduce the impact of low-quality workers and calculate reliable consensus labels.

\noindent  {\adfhalfrightarrowhead} \textbf{\textit{Explanatory annotations}} provide a hard or soft label along with an explanation for each annotation.
For example, 
Schroder et al.~\cite{SchroderBANMWBM20} use topic-related annotations for environmental texts. 
Similarly, Yan et al.\cite{DBLP:conf/aaai/YanHCLX20} annotate the text and list keywords as evidence of the accuracy of the label.
Unlike the above methods, Zhou et al.~\cite{DBLP:conf/aaai/ZhouC0GY21} annotate samples by minimizing correlations between tasks and provide explainable medical knowledge to distinguish selected samples.

\noindent {\adfhalfrightarrowhead} \textbf{\textit{Random/multi-agent annotations}} use multiple independent pseudo-annotators to randomly label new unlabeled samples without human input~\cite{DBLP:journals/pr/YangL19a}. 
For example,
Gong et al.~\cite{DBLP:conf/cvpr/GongFKRL22} use an agent team to collaboratively select informative images for annotation based on the decisions {from} the other agents.

\subsection{\textbf{Query Strategy}}
\label{Query_Strategy}

\noindent {\adfhalfrightarrowhead}  \textbf{\textit{Uncertainty-based methods}} aim to select the most ambiguous samples according to model predictions. Given an input $\bm{x}_{i}$:
\begin{equation}
\text{Entropy}(\bm{x}_{i}) = \mathop{\arg\max}\limits_{\bm{x}_{i}}(\sum_{j} P(\hat{y}_{j}|\bm{x}_{i})\log P(\hat{y}_{j}|\bm{x}_{i})),
\end{equation}
where $P(\hat{y}_{i}|\bm{x}_{i})$ represents the likelihood that $\bm{x}_{i}$ is classified into the $i$-th class~\cite{DBLP:conf/aaai/XieYLLCW22}.
Uncertainty-based methods focus on designing various score functions to measure sample uncertainty and informativeness, including predictive entropy~\cite{DBLP:conf/aaai/XieYLLCW22}, 
least confidence~\cite{DBLP:conf/ijcnn/WangS14}, 
highest estimated dual variables~\cite{DBLP:conf/nips/ElenterNR22},
mutual information between model posterior and predictions~\cite{DBLP:conf/eccv/KothawadeGSXI22}.
Some strategies check samples near the decision boundary as the most uncertain ones~\cite{DBLP:conf/nips/LiDRB20}, such as instances close to the hyperplane~\cite{DBLP:journals/tnn/CarbonneauGG19} or close to the margin~\cite{DBLP:conf/ecml/RothS06}. 
Others combine multiple query strategies, forming a query-by-committee~\cite{BrantleyDS20} or disagreement-based~\cite{DBLP:conf/nips/YanCJ19} DAL strategy to decrease errors made by a single query strategy.
With the development of adversarial learning, instead of selecting samples from unlabeled datasets, models tend to generate the most informative and uncertain synthetic samples to expand the training dataset~\cite{DBLP:conf/icml/TranDRC19}.

However, they have some common drawbacks:
(1) redundant samples, as uncertain points, are continually selected yet in short of coverage;
(2) simply focusing on a single sample lacks robustness to outliers;
(3) these task-specific designs exhibit limited generalizability.

\noindent {\adfhalfrightarrowhead} \textbf{\textit{Representative-based methods}}
aim to sample the most prototypical data points that effectively cover the distribution of the entire feature space.
Existing methods can be categorized into density-based and diversity-based approaches.
\textbf{Density-based} methods prefer to select samples that can represent all unlabeled samples.
They use clustering methods to select cluster centers~\cite{DBLP:journals/pr/YangL22} as the most representative samples or select samples that can maximize probability coverage of the whole feature space of unlabeled datasets~\cite{sener2018active}.
For example, 
Kim et al.~\cite{10.1145/3534678.3539476} design the density awareness coreset approach to estimate sample densities and preferentially select diverse points from sparse regions. Given the input $\bm{x}_{i}$:
\begin{equation}
    \text{Density}(\bm{x}_{i}) = \frac{1}{k} \sum_{j\in \mathcal{N}(\bm{x}_{i}, k)} \|\bm{x}_{i}-\bm{x}_{j}\|_{2}^{2},
\end{equation}
where $\mathcal{N}(\bm{x}_{i},k)$ represents the $k$-nearest neighbors of $\bm{x}_{i}$~\cite{10.1145/3534678.3539476}.
Coleman et al.~\cite{DBLP:conf/aaai/ColemanCKCBBNSZ22} and Gudovskiy et al.~\cite{DBLP:conf/cvpr/GudovskiyHYT20} achieve efficiency by only considering nearest neighbors rather than all data or matching feature densities with self-supervised methods.
\textbf{Diversity-based} methods prefer to select samples that are different from the labeled samples.
They use context-sensitive methods~\cite{DBLP:journals/pami/HasanPMR20} that take into account the distance between a sample and its surrounding labeled samples to enrich the diversity of the labeled dataset. 
BMAL~\cite{DBLP:journals/pami/ChakrabortyBSPY15} performs DAL for the image labeling problem, where diversity is measured by the KL-divergence of the class probabilities distribution of similar neighboring instances, formulated as:
\begin{equation}
\small
    \text{Divergence}(\bm{x}_{i},\bm{x}_{j})= \sum_{j} P(\hat{y}_{j}|\bm{x}_{i})- P(\hat{y}_{j}|\bm{x}_{j})\log \frac{P(\hat{y}_{j}|\bm{x}_{i})}{P(\hat{y}_{j}|\bm{x}_{j})}.
\end{equation}

Other diversity-based methods tend to train a model, such as adversarial networks~\cite{DBLP:conf/cvpr/KimPKC21}, contrastive networks~\cite{DBLP:journals/kbs/JinYQS22}, hierarchical clustering~\cite{DBLP:journals/tnn/CarbonneauGG19}, and pre-trained models~\cite{DBLP:conf/emnlp/Ein-DorHGSDCDAK20}, to help discriminate labeled and unlabeled sets and select the most different unlabeled samples. 
For example, 
Li et al.~\cite{DBLP:conf/ijcai/LiMKYZW20} explicitly learn a non-linear embedding to select representative samples.
Parvaneh et al.~\cite{DBLP:conf/cvpr/ParvanehATHHS22} explore neighborhoods around unlabeled data by interpolating features with labeled points.
Li et al.~\cite{DBLP:conf/nips/LiP0KZ22} propose an acquisition function that measures mutual information between a batch of queries to encourage diversity.
To further increase label efficiency, Citovsky et al.~\cite{DBLP:conf/nips/CitovskyDGKRRK21} use hierarchical clustering to diversify batches, requiring only 40\% of the labels to achieve the same target performance. 
However, since they use ResNet-101 as their backbone, which contains only 170 MB parameters, more than 20\% labeled samples are required for fine-tuning the model.

However, the aforementioned representative-based methods, which solely focus on sampling diverse samples, are always insensitive to samples that are close to the decision boundary (excluding hybrid methods that jointly consider representative and uncertainty), despite the fact that such samples are probably more important to the prediction model, as suggested by Zhao et al.~\cite{DBLP:journals/pr/ZhaoSZCG19}. 
In addition, representative-based methods work well for a small sample of data and classifiers with a small number of classes since their computational complexity is almost quadratic with respect to data size~\cite{DBLP:conf/nips/CitovskyDGKRRK21}.

\noindent {\adfhalfrightarrowhead} \textbf{\textit{Influence-based methods}} aim to select samples that will have the greatest impact on the performance of the target model.
These techniques can be categorized into three main groups.
(1) The first group is directly measuring the expected impact on the modal through metrics such as gradient norm~\cite{DBLP:conf/aaai/Wang0YHZH0022}, 
query complexity~\cite{DBLP:conf/nips/ThiessenG21},
kernel approximation~\cite{DBLP:conf/nips/MohamadiBS22},
KL divergence~\cite{DBLP:conf/cvpr/GudovskiyHYT20},
change of loss function~\cite{DBLP:conf/cvpr/YooK19}, or model parameters~\cite{DBLP:conf/iccv/HuangWXHD21}, and expected error reduction (EER)~\cite{DBLP:conf/nips/ZhaoDYAQ21}. Specifically, EER can be formulated as 
\begin{equation}
\small
\text{EER}(\bm{x}_{i}) =  \mathbb{E}_{\bm{x}_{s}}\{\mathbb{E}_{y_{i}|\bm{x}_{i}}[\max_{y_{s}} p(y_{s}|\bm{x}_{s},\bm{x}_{i},y_{i})]- \max_{y_{s}} p(y_{s}|\bm{x}_{s})\},
\end{equation}
where $\bm{x}_{s}$ refers to the labeled sample.
(2) The second group is incorporating different learning policies, such as reinforcement learning and imitation learning, to select samples based on reward signals or demonstrated actions.
Despite the promising advantages, this requires significant additional training~\cite{vu-etal-2019-learning}.
For example, Wertz et al.~\cite{DBLP:conf/acl/WertzBMK23} propose reinforced DAL, a reinforcement learning policy that uses multiple elements of the data and the task to dynamically pick the most useful unlabeled subset during the DAL process;
(3) The last group is training a separate model to estimate the impact on the target model~\cite{DBLP:conf/nips/ElenterNR22}.
For example, 
Peng et al.~\cite{DBLP:conf/iccv/PengWLY21} propose a knowledge distillation framework to evaluate the impact of samples based on the knowledge learned by the student model. 
Elenter et al.~\cite{DBLP:conf/nips/ElenterNR22} use the dual variables of the original model to measure the impact on the target model.

However, despite recent advances, influence-based DAL remains challenging. Directly measuring model changes or incorporating new learning policies always requires huge time and space costs, and training a new model will over-rely on its accuracy and often lead to unstable results.

\noindent {\adfhalfrightarrowhead} \textbf{\textit{Bayesian methods}} 
aim to minimize classification errors and improve model beliefs by leveraging Bayes' rule. 
Most studies have treated Bayesian models (e.g., Gaussian process~\cite{DBLP:conf/nips/ZhaoDYAQ21}, BNNs~\cite{DBLP:conf/icml/GalIG17}, Bayesian probabilistic ensemble~\cite{DBLP:conf/cvpr/BeluchGNK18}) as uncertainty-based methods, using them to estimate the informativeness of the sample. 
However, Bayesian DAL is better viewed as its own distinct system, with methods that select batches by directly measuring impact on the target model, such as BatchBALD~\cite{DBLP:conf/nips/KirschAG19} and Causal-BALD~\cite{DBLP:conf/nips/JessonTAKSG21}.
For example, we define a Bayesian model with model parameters $\bm{w} \sim p(\bm{w}|\mathcal{D}_{\text{train}})$, and BALD can be defined to estimate the mutual information between the model predictions and the model parameters, formulated as: 
\begin{equation}
\small
    \mathbb{I}(y;\bm{w}|\bm{x},\mathcal{D}_{\text{train}}) = \mathbb{H}(y|\bm{x},\mathcal{D}_{\text{train}}) - \mathbb{E}_{p(\bm{w}|\mathcal{D}_{\text{train}})}[\mathbb{H}(y|\bm{x},\bm{w},\mathcal{D}_{\text{train}})],
\end{equation}
where $\mathbb{H}$ represents the entropy and $\mathbb{E}$ is the expectation.

Compared to standard DNNs, the aforementioned Bayesian DAL methods, which leverage the advantages of probabilistic graphical theory~\cite{DBLP:conf/icml/GalIG17}, can often provide reasonable explanations for why these samples should be selected~\cite{DBLP:conf/nips/KirschAG19}. However, they often require extensive accurate prior knowledge and tend to underperform deep learning models in representation learning and fitting capacity. 

\noindent {\adfhalfrightarrowhead} \textbf{\textit{Hybrid methods}} aim to take advantage of the above multiple query strategies and to achieve a trade-off among them. 
Hybrid methods can be further categorized according to interaction patterns.
\textbf{Serial-form hybrids} apply criteria sequentially within an DAL cycle, filtering out non-informative samples until the batch is filled~\cite{DBLP:conf/nips/CitovskyDGKRRK21}. 
\textbf{Criteria-selection hybrids} use only one query strategy in one DAL iteration, in which they select the best query strategy or network architecture with the highest criterion. 
For example,
DUAL~\cite{DBLP:conf/ecml/DonmezCB07} switches between density-based and uncertainty-based selectors to choose the best criterion for each DAL cycle.
Unlike DUAL, iNAS~\cite{DBLP:conf/nips/GeifmanE19} searches a restricted candidate set to find the optimal model architecture incrementally in each DAL iteration.
\textbf{Parallel-form hybrids} use multi-objective optimization methods or a weighted sum to merge multiple query criteria into one for sample selection. 
For example,
Gu et al.~\cite{DBLP:journals/tnn/GuZDH21} efficiently acquire batches with discriminative and representative samples by proposing procedures to update labeled and unlabeled sets, based on path-following optimization techniques.
Citovsky et al.~\cite{DBLP:conf/nips/CitovskyDGKRRK21} jointly optimize the uncertainty and diversity criteria in batch mode  using multi-objective acquisition functions.
TOD~\cite{DBLP:conf/iccv/HuangWXHD21} selects samples with high model uncertainty and outputs discrepancy through a weighted combination of both metrics.

Hybrid methods combine the advantages of different query strategies. 
However, determining the most effective combinations and trade-offs between criteria is time consuming and still remains open for further investigation.

\subsection{\textbf{Model Architecture}}
\label{deep_arch}


\noindent  {\adfhalfrightarrowhead} \textbf{\textit{Traditional Machine Learning}} architectures, such as Forest~\cite{DBLP:conf/nips/KonyushkovaSF17} and 
Support Vector Machine (SVM)~\cite{DBLP:journals/tnn/CarbonneauGG19}, are statistical-based models that do not use neural networks. And they attract great attention in the early stage of the DAL development.

\noindent  {\adfhalfrightarrowhead} \textbf{\textit{Bayesian Neural Networks}} (BNNs) combine neural networks with Bayesian inference, quantifying the uncertainty introduced by the models in terms of outputs and weights to explain the trustworthiness of the prediction~\cite{DBLP:conf/icml/IzmailovVHW21}.
Many studies propose DAL strategies based on BNNs, aiming to improve efficiency and explainability in samples selection~\cite{DBLP:conf/emnlp/FangLC17,DBLP:conf/nips/KirschAG19}.

\noindent  {\adfhalfrightarrowhead} \textbf{\textit{Recurrent Neural Networks}} (RNNs)~\cite{rumelhart1986learning} use their reasoning from previous experiences to predict upcoming events and are able to learn features with long-term dependencies. 
They have been widely used for sequential data such as text and audio.
DAL is seldom combined with RNNs since they require large-scale labeled datasets for training. 
Some special tasks that easily recognizable patterns, such as malicious word detection on social networks~\cite{DBLP:conf/acl-deeplo/ZengGCNP19}, can be solved with DAL.

\noindent  {\adfhalfrightarrowhead} \textbf{\textit{Convolutional Neural Networks}} (CNNs)~\cite{DBLP:journals/pieee/LeCunBBH98} are feedforward neural networks that can extract features from data with convolution structures and have been widely used for image processing {with} three advantages: local connections, weight sharing, and down-sampling dimensionality reduction.
DAL can be effectively combined with CNNs since Sener et al.~\cite{sener2018active} proved that a subset of samples (coreset) can geometrically characterize all features of the entire image set and can be selected by minimizing a rigorous bound.
Following their study, more studies have been
conducted~\cite{DBLP:conf/iccv/SinhaED19,DBLP:conf/nips/CitovskyDGKRRK21}.

\noindent  {\adfhalfrightarrowhead} \textbf{\textit{Graph Neural Networks}} (GNNs)~\cite{DBLP:conf/iclr/KipfW17} learn node representations by aggregating neighborhood information and achieve great success in various tasks, such as node classification.
However, effectively handling graph data with dense interconnections between samples using limited labeled data remains an open challenge~\cite{DBLP:journals/nn/AmirinezhadSH22}.
DAL can help address this by selectively querying labels for the most informative samples and executing only one training epoch to reduce the annotation cost for various types of graphs, such as homogeneous graphs~\cite{DBLP:conf/nips/HuXQYC0T20}, heterogeneous graphs~\cite{DBLP:conf/icdm/RenWZ020}
and attribute graphs~\cite{DBLP:journals/tnn/LiYC21}.

\noindent  {\adfhalfrightarrowhead} \textbf{\textit{Variational Autoencoders}} (VAEs) is a class of neural network architecture designed with an encoder-decoder framework~\cite{DBLP:journals/corr/KingmaW13}. 
It aims to capture the underlying data distribution and learn to generate samples that closely resemble the input data.
VAEs-based DAL methods usually generate samples to fool discriminators in an adversarial training manner, 
thus improving discriminators' ability to select the most challenging-to-distinguish samples for training DAL models~\cite{DBLP:conf/iccv/SinhaED19,DBLP:conf/cvpr/KimPKC21}.

\noindent  {\adfhalfrightarrowhead} \textbf{\textit{Pre-Trained Language Models (PLMs)}}, based on Transformers, utilize multi-head self-attention to capture long-term dependencies.
By pre-training on large unlabeled corpora, PLMs embed substantial general knowledge and transfer to downstream tasks, enabling state-of-the-art (SOTA) performance~\cite{ayub2022fewshot}. 
For example, Seo et al.~\cite{DBLP:conf/aaai/SeoKAL22} identify the most informative samples for a given task, focusing on PLMs fine-tuning, to learn salient patterns with minimal annotation cost. 
The combination of pre-training rich knowledge foundation and DAL's sample-efficient tuning unlocks PLMs 's further potential for many applications.

\subsection{\textbf{Learning Paradigm}}
\label{Learning_Paradigm}

\noindent {\adfhalfrightarrowhead} \textbf{\textit{Traditional Learning Paradigm}}, as illustrated in Algorithm~\ref{deep_algorithm}, iteratively queries and labels samples to train the models in a vanilla supervised learning manner, without incorporating any advanced learning paradigms~\cite{DBLP:journals/corr/GalG15a,DBLP:conf/aaai/SeoKAL22}.

\noindent {\adfhalfrightarrowhead} \textbf{\textit{Semi-supervised Learning}}, also known as weakly-supervised learning, aims to jointly use real-labeled samples and pseudo-labeled samples to train the models.
Current DAL methods are designed with various efficient strategies to obtain pseudo-labels for unlabeled samples.
For instance, DBAL~\cite{DBLP:conf/icml/GalIG17} and CoreSet~\cite{sener2018active} first predict pseudo-labels using their models and then calculate samples' confidence scores to judge whether these pseudo-labels should be trusted or not.  On the other hand, LADA~\cite{DBLP:conf/nips/KimSJM21} and BGADL~\cite{DBLP:conf/icml/TranDRC19} propose new data augmentation methods to create more samples based on original labeled samples, using their original real-labeled samples as pseudo-labels.
These studies effectively reduce human-labors and achieve comparable performance compared with traditional supervised learning using larger labeled samples.

\begin{figure}[t]
\includegraphics[width=0.48\textwidth]{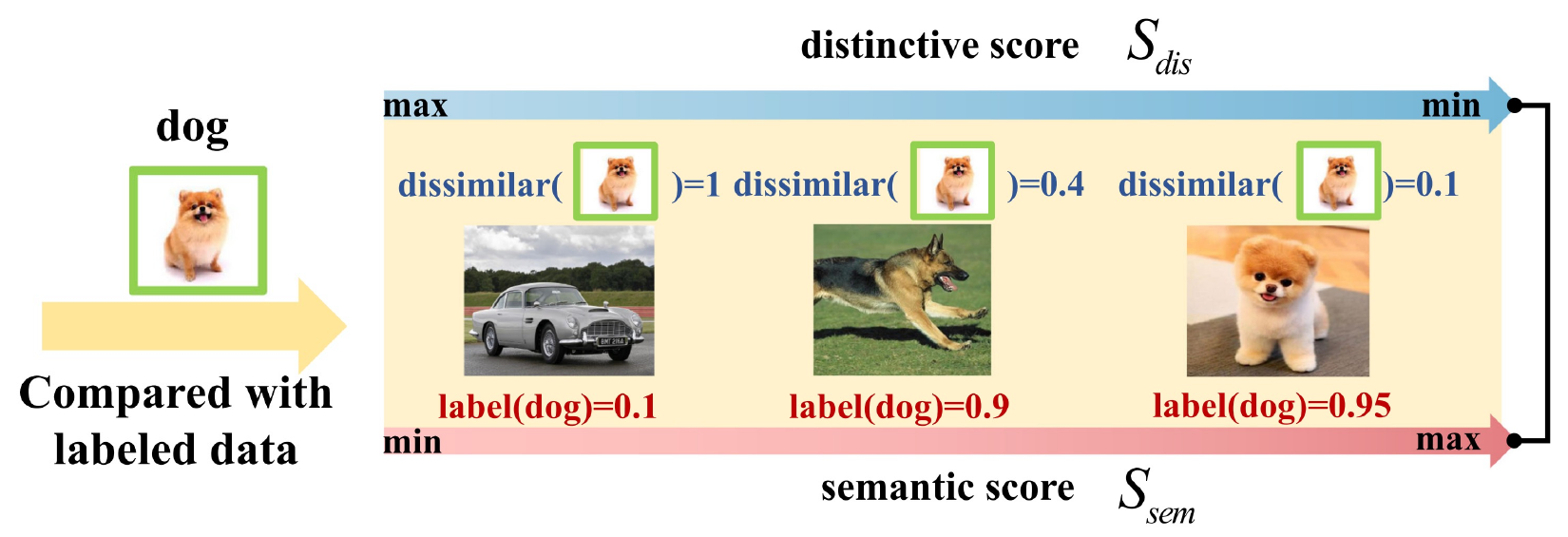}
\caption{An example for contrastive learning based query strategies.}
\label{pipeline-CL}

\end{figure}

\noindent  {\adfhalfrightarrowhead} \textbf{\textit{Contrastive Learning}} improves feature representation by pulling similar instances closer together while pushing dissimilar instances apart~\cite{DBLP:journals/entropy/Albelwi22}.
Contrastive methods extract discriminative features, such as semantics~\cite{DBLP:journals/kbs/JinYQS22} and distinctiveness~\cite{DBLP:conf/cvpr/KimPKC21}, to estimate the sample uncertainty during acquisition.
For example, as shown in Fig.~\ref{pipeline-CL},
Du et al.~\cite{DBLP:journals/pami/DuCZCCL23} extract both semantic and distinctive features with contrastive learning and then combine them in a query strategy to choose the most informative unlabeled samples with matched categories.

\begin{figure}[h]
\includegraphics[width=0.47\textwidth]{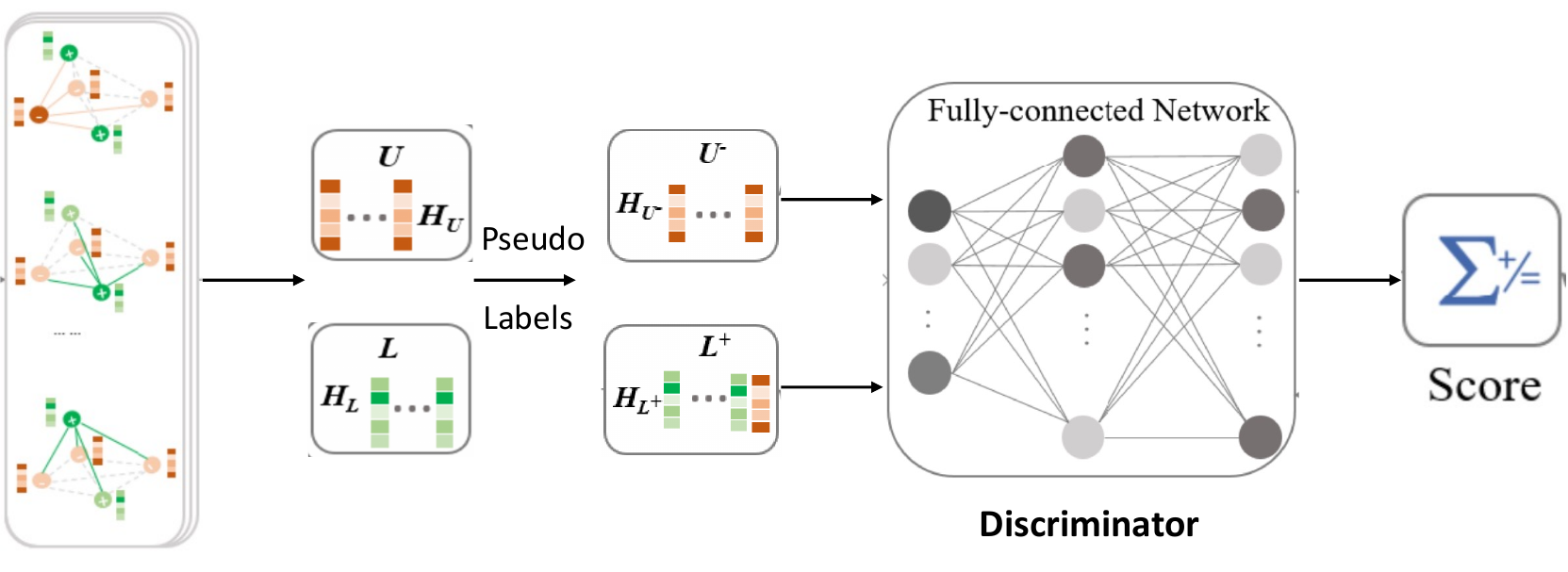}
\caption{An example for contrastive learning based query strategies.}
\label{pipeline-AD}
\end{figure}

\noindent  {\adfhalfrightarrowhead} \textbf{\textit{Adversarial Learning}} enables a model to train fully differentiable by solving minimax optimization problems~\cite{DBLP:conf/iccv/SinhaED19}.
This approach can be used as a generative query technique for DAL. 
For example, DAL can be combined with generative adversarial network, which consist of a generator and a discriminator, where the DAL model acts as the discriminator and the generator explores the distribution of unlabeled data to generate the most informative and uncertain synthetic samples for training~\cite{DBLP:conf/cvpr/KimPKC21}.
Li et al.~\cite{DBLP:journals/tnn/LiYC21} propose SEAL, as shown in Fig.~\ref{pipeline-AD} which consists of two adversarial components. The graph embedding network encodes all nodes into a shared space, with the intention of making the discriminator treat all nodes as labeled. Additionally, a semi-supervised discriminator is used to differentiate unlabeled nodes from labeled ones. The divergence score of the discriminator is used as an informativeness measure to actively select the most informative node for labeling. The two components form a loop to mutually improve DAL.

\noindent {\adfhalfrightarrowhead} \textbf{\textit{Meta Learning}} enables DNNs to leverage the knowledge acquired from multiple tasks, represented in the network with their weights, to adapt faster to new tasks.
Meta learning can provide an acquisition function for DAL~\cite{DBLP:conf/nips/KonyushkovaSF17,DBLP:conf/nips/ParkSBLS022} or favorable model initialization during DAL by controlling the transfer of knowledge from multiple source tasks.
For example, Shao et al.~\cite{DBLP:conf/icdm/ShaoWL19} propose Learning-to-Sample, where a boosting model and sampling model dynamically learn from each other and iteratively improve performance. Zhu et al.~\cite{zilongzhu} combine both paradigms by initializing an active learner with meta-learned parameters via meta-training on tasks similar to the target task.

\noindent  {\adfhalfrightarrowhead} \textbf{\textit{Reinforcement Learning}} involves an agent that can interact with its environment and learn to alter its behavior in response to received rewards~\cite{DBLP:journals/nn/AmirinezhadSH22}. 
Given that almost all DAL methods use heuristic acquisition functions with limited effectiveness, 
Reinforcement learning frames DAL as a reinforcement learning problem to explicitly optimize an acquisition policy. 
In the DAL with reinforcement learning setup, an autonomous agent (acquisition selector) controlled by a deep learning algorithm that observes a state $s_{t}$ from its environment (predictor) at time {$t$}. 
It takes an action $a_{t}$ to maximize the reward $r_{t}$ (prediction accuracy), where $a_{t}$ decides whether to query unlabeled samples~\cite{DBLP:conf/icml/MenardDJKLV21}.

\noindent {\adfhalfrightarrowhead} \textbf{\textit{Curriculum Learning}} mimic human and animal learning processes, where the training progresses gradually from simple to complex samples. This provides a natural way to exploit labeled data for robust learning~\cite{DBLP:conf/icml/BengioLCW09,DBLP:conf/aaai/JiangMZSH15}.
Specifically, curriculum learning uses a predefined learning constraint to incrementally incorporate additional labeled samples during training. 
Curriculum Learning introduces a weighted loss on all labeled samples, acting as a general regularizer over the sample weights. 
For example,
Wang et al.~\cite{DBLP:journals/pami/LinWMZZ18} use a pseudo-labels strategy which iteratively assigns pseudo-labels to unlabeled samples with high prediction confidence.

\noindent {\adfhalfrightarrowhead} \textbf{\textit{{Continual Learning}}} is developed for constraints on task-based settings, where the model continuously learns a sequence of tasks one at a time, where all data for the current task are labeled and available in increments.
However, real-world systems do not have the luxury of large labeled datasets for each new task. 
To address this issue,
Mundt et al.~\cite{DBLP:journals/nn/MundtHPR23} present a detailed analysis of continual learning-based DAL and out-of-distribution detection works. They suggest a unified perspective with open-set recognition as a natural interface between continual learning and DAL.
Ayub et al.~\cite{ayub2022fewshot} develop a method that allows a continue learning agent to continually learn new object classes from a few labeled examples.

\begin{figure}[h]
\includegraphics[width=0.48\textwidth]{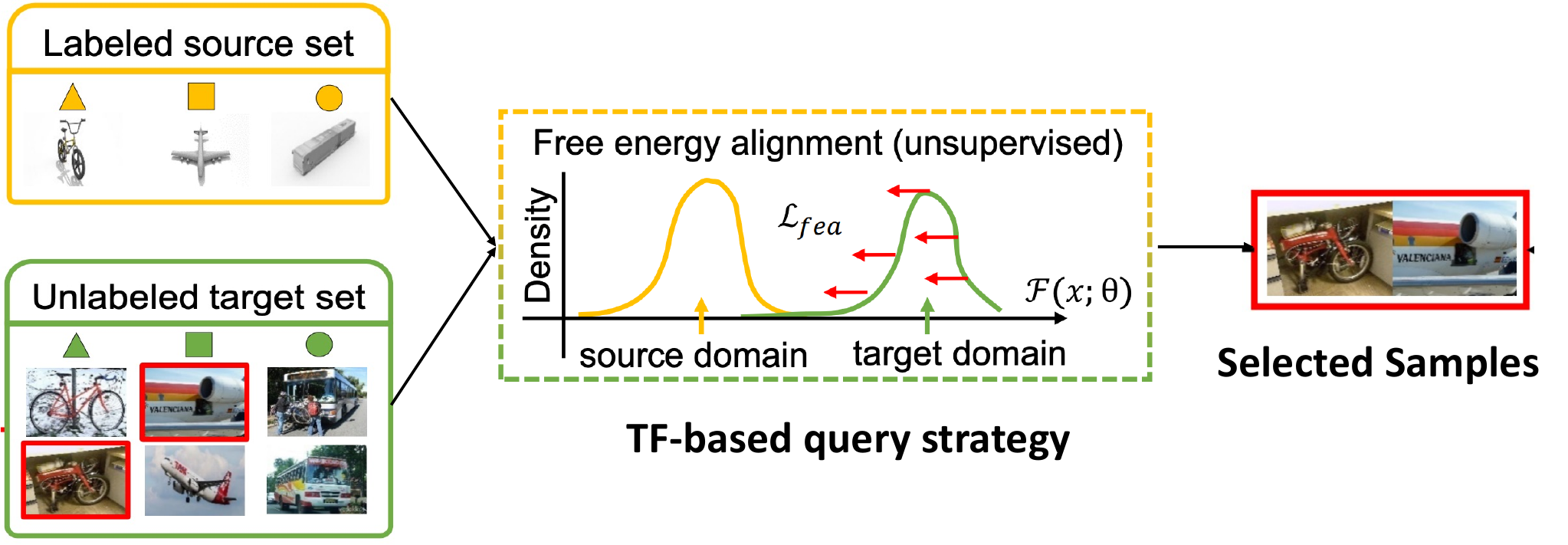}
\caption{An example for transfer learning based query strategies.}
\label{pipeline-TF}
\end{figure}

\noindent {\adfhalfrightarrowhead} \textbf{\textit{Transfer Learning}} extracts knowledge from one or more source tasks and applies it to a target task. It has two broad categories: transductive and inductive. 
While transductive methods adapt models learned from a labeled source domain to a different unlabeled target domain with the same task,
inductive methods ensure that the domains of source and target are the same but tasks are different. 
DAL with transfer learning can better enhance each other's performance by selecting the best target samples with a distribution similar to the source domain~\cite{DBLP:conf/wacv/SuTSLMC20}. 
In addition, transfer learning can minimize the number of annotation labels needed and provide auxiliary information for DAL acquisition functions. 
For example, as shown in Fig.~\ref{pipeline-TF}, Xie et al.~\cite{DBLP:conf/aaai/XieYLLCW22} propose an energy-based active domain adaptation that balances domain representation and uncertainty when selecting target data.

\begin{figure}[h]
\includegraphics[width=0.48\textwidth]{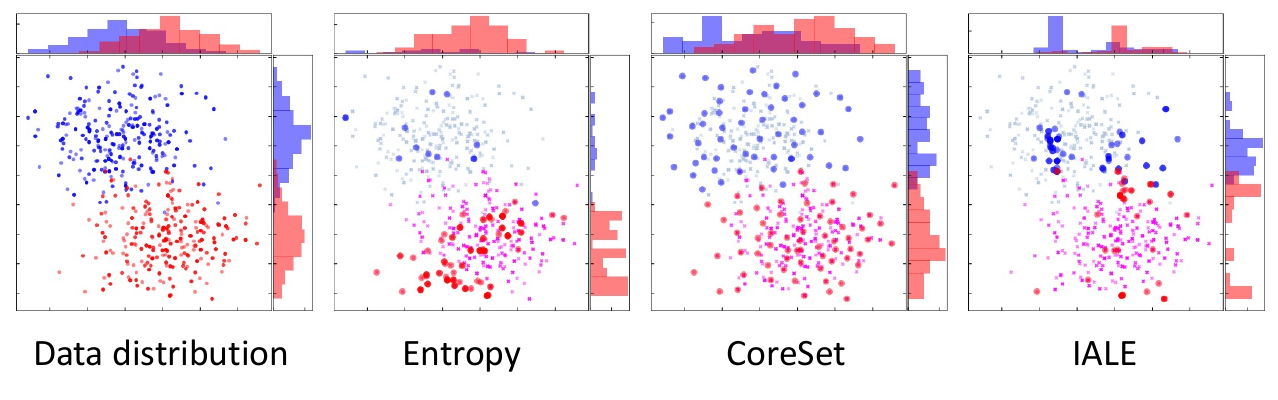}
\caption{An example for imitation learning~\cite{DBLP:journals/jmlr/LofflerM22}.}
\label{pipeline-IM}
\end{figure}

\noindent {\adfhalfrightarrowhead} \textbf{\textit{Imitation Learning}} provides SOTA results in many structured prediction tasks by learning near-optimal search policies~\cite{BrantleyDS20}.
Such methods assume access to an expert during training that can provide the optimal action in any queried state, essentially asking ``what would you do here?'' and learning to mimic that choice.
For example, 
Bullard et al.~\cite{DBLP:conf/ijcai/BullardSC19} use imitation learning to allow an agent in a constrained environment to concurrently reason about both its internal learning goals and externally impose environmental constraints within its objective function.
Löffler et al.~\cite{DBLP:journals/jmlr/LofflerM22} propose an imitation learning scheme (IALE) that mimics the selection of the best-performing expert heuristic at each stage of the learning cycle in a batch-mode setting.
As shown in Fig.~\ref{pipeline-IM}, IALE can well imitate the Entropy-based and CoreSet-based methods and thus obtain better performance.

\noindent {\adfhalfrightarrowhead} \textbf{\textit{Multi-task Learning}} (MTL) focuses on formulating methods to maintain performance across multiple tasks rather than a single task. 
Multi-task DAL (MTAL) methods combine multiple individual task-related query strategies into a single unified approach and jointly optimize the unified one.
In contrast to single-task query settings, where the uncertainty of a single selected task classifier is used to query unlabeled samples, in MTAL the uncertainty of an instance is determined by the uncertainties from classifiers across all tasks.
For example, 
Ikhwantri et al.~\cite{ikhwantri-etal-2018-multi} propose an MTAL framework for semantic role labeling with entity recognition as an auxiliary task.
This alleviated data needs and leverages entity information to aid role labeling. Their experiments show that MTAL can outperform single-task DAL and standard MTL, using 12\% less training data than passive learning.
Zhou et al.~\cite{DBLP:conf/aaai/ZhouC0GY21} propose a Multi-Task Adversarial DAL framework, where adversarial learning maintains the effectiveness of the MTL and DAL modules. A task discriminator eliminates irregular task-specific features, while a diversity discriminator exploits heterogeneity between samples to satisfy diversity constraints.

\subsection{\textbf{Training Process}}
\label{Training_Process}

\noindent  {\adfhalfrightarrowhead} \textbf{\textit{Traditional Training}}
first trains a model on an initialized training dataset and then selects unlabeled samples to annotate based on the predictions of the current model. 
The newly annotated samples are added to the training set for re-training the model in the next iteration~\cite{DBLP:conf/ijcai/HuangZNY21}. 
This iterative process continues, with the model parameters randomly re-initialized before each epoch of re-training~\cite{DBLP:journals/tcsv/WangZLZL17}, until either the sample budget or number of DAL iterations is reached.

\noindent {\adfhalfrightarrowhead} \textbf{\textit{Curriculum Learning Training}} 
gradually progresses from easy to complex samples, mimicking human and animal learning processes. 
This provides a natural and iterative way to exploit labeled data for robust learning. 
For example,  Tang et al.~\cite{DBLP:conf/aaai/TangH19} propose a self-paced DAL approach that jointly considers the value and difficulty of a sample. It queries samples from easy to hard to minimize annotation cost. 
Wang et al.~\cite{DBLP:journals/tnn/WangLYCZZ19} show that curriculum learning alone improves the accuracy of the object detection by 3.6\%, while the combination of curriculum learning and DAL improve the accuracy by 4.3\%.

\noindent  {\adfhalfrightarrowhead} \textbf{\textit{Pre-training \& Fine-tuning (Pre+FT)}} have become a primary training process with the development of large-scale PLMs~\cite{Outliers-ACL1}.  
It leverages the rich prior knowledge in PLMs to solve different downstream tasks.
DAL attracts attention as a sample selection strategy for fine-tuning with only 10\%$\sim$20\% of labeled data achieving competitive performance compared to full data fine-tuning~\cite{DBLP:conf/aaai/SeoKAL22}.
DAL iteratively selects and annotates batches of informative samples to fine-tune the PLMs for the downstream task. This satisfies task-specific needs, while also enabling a few-shot learning~\cite{ayub2022fewshot}.

\begin{table*}[t]
\centering
\scriptsize
\caption{Illustration of DAL-related applications in main fields,  including classic methods with their advantages and disadvantages.}
\setlength{\tabcolsep}{1.6mm}{
\begin{tabular}{@{}cllll@{}}
\toprule
Areas                     
& \multicolumn{1}{l}{Applications}                
& \multicolumn{1}{c}{Classic Methods}                                   
& \multicolumn{1}{c}{Advantages}  
& \multicolumn{1}{c}{Disadvantages}          
\\ 
\midrule
\multirow{16}{*}{NLP}                   
& \multirow{3}{*}{Text Classification}    
& generate samples for training~\cite{DBLP:conf/aaai/YanHCLX20, DBLP:conf/nips/TanDB21}.             
& make the selection process efficient.                                
& high time consumption, unstable performance.                
\\
&          
& uncertainty sampling~\cite{DBLP:conf/acl/SchroderNP22}.            
& high efficiency and performance.                         
& vulnerable to outliers, unstable performance.                  \\
&          
& use pre-trained language models~\cite{DBLP:conf/acl/JelenicJDS23}.              
& easily adapt to new datasets.   
& vulnerable to outliers and imbalanced datasets.                               \\
\specialrule{0em}{1pt}{1pt}
\cmidrule(l){2-5}
& \multirow{2}{*}{Text Summarization} 
& PLMs with Monte Carlo dropout~\cite{GidiotisT22}.
& efficient and effectiveness.    
&  vulnerable to outliers, unstable performance.                                   \\
&          
& diverse sampling~\cite{TsvigunLSLDKBSP22}.                            
& remove outliers and diverse sampling.                    
& vulnerable to document embeddings. \\
\specialrule{0em}{1pt}{1pt}
\cmidrule(l){2-5}
& \multirow{2}{*}{Question Answering} 
& DataMap~\cite{Outliers-ACL1}.                                 
& eliminate outliers and improve accuracy.      
& high time consumption, lack of generalizability.                 \\
&          
& interactive query strategy~\cite{DBLP:conf/aaai/PadmakumarM21}.                 
& efficiently minimize costly data annotations.                  
& wait for human reaction, need expert knowledge.
\\
\specialrule{0em}{1pt}{1pt}
\cmidrule(l){2-5}
& \multirow{2}{*}{Information Extraction}         
& label identical subsequences~\cite{RadmardFL20}          
& high efficiency and effectiveness.                         
& lack of generalizability, cold-start.                             \\
&          
& label most novel words~\cite{Hua022}.                   
& high efficiency and effectiveness.                  
& unstable performance, cold-start.          
\\
\specialrule{0em}{1pt}{1pt}
\cmidrule(l){2-5}
& \multirow{2}{*}{Semantic Parsing}               
& hyperparameter selection~\cite{DuongAEPCJ18}.                     
& reduce data annotation.                            
& high time consumption, lack of generalizability.        \\
&          
& hybrid query strategies~\cite{DBLP:conf/acl/LiQCTH23}.                      
& select the most semantically varied samples.                  
& vulnerable to outliers, lack of scalability.              \\
\midrule
\multirow{14.5}{*}{CV}      
& \multirow{2}{*}{Image Captioning}               
& semantic adversarial DAL~\cite{DBLP:conf/mm/ZhangL0WDZH20}         
& overcome scarcity of labeled data.              
& difficulty in cross-domain transfer, cold-start.                        \\
&          
& domain transfer learning~\cite{DBLP:conf/lion/CheikhZ20}.
& transfer knowledge from high-resource.                    
& vulnerable to outliers, data scarcity.                      \\
\specialrule{0em}{1pt}{1pt}
\cmidrule(l){2-5}
& \multirow{2}{*}{Semantic Segmentation}          
& uncertainty-based DAL~\cite{DBLP:journals/cviu/KonyushkovaSF19}.                   
& high efficiency and effectiveness.                   
& unstable performance,  easily select outliers.                                    \\
&          
& region-based selection~\cite{DBLP:conf/aaai/QiaoZLZWDY22,DBLP:conf/cvpr/XieY0LC22}.
& balance between label efforts and effect.           
& vulnerable to outliers, imbalance datasets.               \\
\specialrule{0em}{1pt}{1pt}
\cmidrule(l){2-5}
& \multirow{2}{*}{Object Detection}               
& hybrid selection~\cite{DBLP:conf/cvpr/WuC022,DBLP:journals/tnn/WangLYCZZ19}.
& avoid noisy samples and outliers.                     
& data scaricity, unstable performance.      
\\
&         
& instance uncertainty learning~\cite{DBLP:conf/cvpr/YuanWFLXJY21}.                
& suppress noisy instances.        
& unstable performance, lack of scalability.                                       \\
\specialrule{0em}{1pt}{1pt}
\cmidrule(l){2-5}
& \multirow{2}{*}{Pose Estimation}                
& traditional DAL strategy~\cite{DBLP:conf/wacv/CaramalauBK21,DBLP:conf/cvpr/ShuklaA21}. 
& effectiveness, easy to apply.         
& vulnerable to outliers, cold-start.                          \\
&          
& meta learning~\cite{DBLP:conf/cvpr/GongFKRL22}.                              
& can learn an optimal sampling policy.       
& vulnerable to outliers and imbalance datasets.                                     \\
\specialrule{0em}{1pt}{1pt}
\cmidrule(l){2-5}
& \multirow{2}{*}{Target Tracking}                
& multi-frame collaboration~\cite{DBLP:journals/corr/abs-2110-13259}                 
& eliminate background noise, ensure diversity.              
& unstable performance, lack of scalability.                   \\
&          
& multi-target object tracking~\cite{DBLP:journals/corr/abs-2205-03555}.        
& high efficient and effectiveness.                  
& high time consumption, cold-start                                 \\
\specialrule{0em}{1pt}{1pt}
\cmidrule(l){2-5}
& \multirow{2}{*}{Person Re-identification}       
& human-in-the-loop~\cite{DBLP:conf/iccv/LiuWGTL19}.                  
& improve model performance.      
& high time consumption, lack of generalizability.     \\
&         
& incremental annotation~\cite{DBLP:journals/pr/XuLZGH21}.                      
& select diverse samples without redundancy.             
& vulnerable to outliers, cold-start.                          \\
\midrule
\multicolumn{1}{l}{\multirow{7}{*}{DM}} 
& \multirow{2}{*}{Node Classification}            
& semi-supervised adversatial DAL~\cite{DBLP:journals/tnn/LiYC21}.              
& better performance gains.         
& unstable performance, cold-start.          
\\
\multicolumn{1}{l}{}                    
&          
& graph policy network~\cite{DBLP:conf/nips/HuXQYC0T20}.                       
& stable performance.                 
& single sample selection costs much time.                \\
\specialrule{0em}{1pt}{1pt}
\cmidrule(l){2-5}
\multicolumn{1}{l}{}                 
& \multirow{2}{*}{Link Prediction}   
& multi-view DAL~\cite{DBLP:conf/ijcai/CaiTCHWH19}.                      
& query informative samples from multi-view.      
& lack of scalability and generalizability.                     \\
\multicolumn{1}{l}{}                    
&          
& transfer learning DAL~\cite{DBLP:journals/ai/ZhaoPY17}.                          
& easily apply to new datasets.                   
& unstable performance, cold-start.          
\\
\specialrule{0em}{1pt}{1pt}
\cmidrule(l){2-5}
\multicolumn{1}{l}{}                  
& \multirow{2}{*}{Community Detection}            
& topic-based~\cite{DBLP:journals/grid/GuptaJS18}.                                
& reducing the unreliable dataset. 
& high time consumption, unstable performance.\\
\multicolumn{1}{l}{}                    
&          
& geometric block model~\cite{DBLP:conf/aaai/ChienTL20}.                       
& efficient and effectiveness.           
& unstable performance, cold-start.          \\ 
\bottomrule
\end{tabular}}
\label{table: DAL-related applications}
\end{table*}

\section{Applications of DAL}
\label{Application}

As shown in Table~\ref{table: DAL-related applications}, the integration of DL and AL is leading to an increasing application of AL methods in various domains of life, ranging from agricultural development~\cite{SchroderBANMWBM20} to industrial revitalization~\cite{SchroderBANMWBM20}, and from artificial intelligence~\cite{DBLP:conf/acl/JelenicJDS23} to biomedical fields~\cite{DBLP:conf/adprl/DengPM11}. 
In this section, we aim to provide a systematic and detailed overview of existing DAL-related work from a broad application perspective.

\subsection{\textbf{Applications in Natural Language Processing}}

With the emergence of large-scale language models, NLP has achieved great success using computers to help understand intricate languages.
However, fine-tuning these language models requires a substantial amount of data, computation resources, and time. DAL provides a strategy for searching high-quality small and high-quality samples to help fine-tune the model and save resources.
In the following, we introduce some of the most influential DAL methods in NLP.

\noindent {\adfhalfrightarrowhead} \textbf{\textit{Text Classification}} aims to classify large-scale text with particular labels such as topic or sentiment. 
Researchers propose several methods to efficiently select informative samples for training. 
For example,
Yan et al.~\cite{DBLP:conf/aaai/YanHCLX20}  generate the most informative examples for training, efficiently skipping the sample selection process. 
They approximate the generated example with a few summary words, which significantly reduces the labeling cost for annotators, as they only need to read a few words instead of a long document.
Tan et al.~\cite{DBLP:conf/nips/TanDB21} develop the Bayesian estimate of mean proper scores (BEMPS) framework for DAL, which allows the calculation of scores such as logarithmic probability to better help select informative and uncertainty samples. 
Experiments demonstrate that BEMPS is more effective than baselines in various text classification datasets.
On the other hand, Schroder et al.~\cite{DBLP:conf/acl/SchroderNP22} use transformers for uncertainty-based sample selection. Interestingly, they achieve comparable performance in widely used text classification datasets while training in less than 20\% of the labeled data, which demonstrates their ability to utilize limited labeled data.
In another study, Jelenic et al.~\cite{DBLP:conf/acl/JelenicJDS23} conduct an initial empirical study to investigate the transferability of the DAL by using PLMs . They find DAL can effectively adapt to new datasets with pre-trained models.

\begin{figure}[t]
\centering
\includegraphics[width=0.47\textwidth]{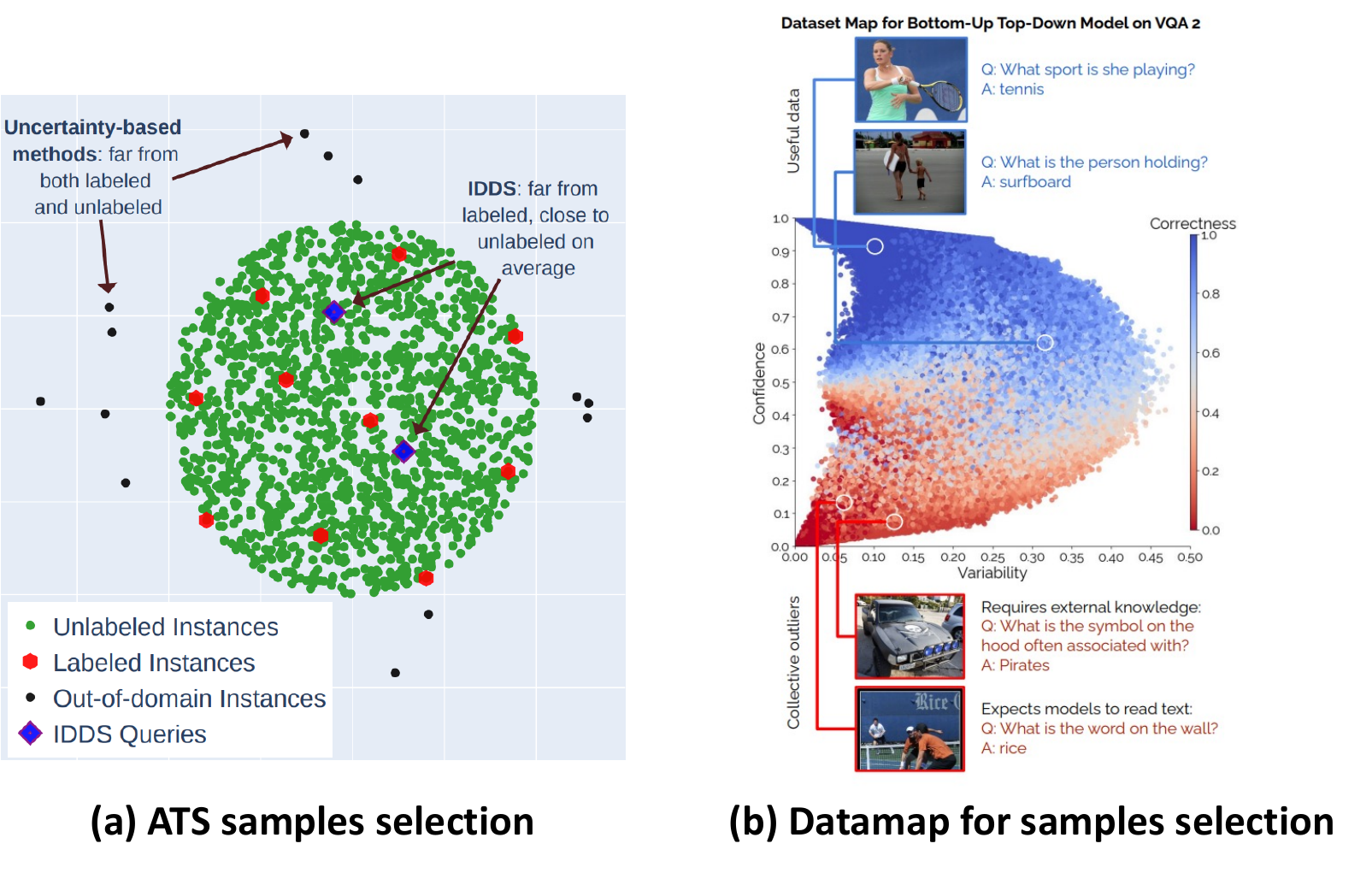}
\caption{An example for samples selection of ATS and Datamap.}
\label{pipeline-ATS}

\end{figure}

\noindent {\adfhalfrightarrowhead} \textbf{\textit{Abstractive Text Summarization} (ATS)}
\label{summarization}
aims to compress a document into a brief, informative and readable summary that retains the key information of the original document.
However, constructing human-annotated datasets is a time-consuming and costly endeavor. 
DAL are explored to reduce the amount of annotation needed while achieving a certain level of ATS performance.
For example, 
Gidiotis et al.~\cite{GidiotisT22} address the issue from a Bayesian view and study uncertainty estimation for SOTA text summarization models. They augment the pre-trained summarization models with Monte Carlo dropout, forming the corresponding variational Bayesian PLMs models. By generating multiple summaries from these models, they approximate Bayesian inference and estimate the summarization uncertainty. 
Experiments on multiple benchmark datasets consistently demonstrate their improved summarization performance with higher Recall-Oriented Understudy for Gisting Evaluation (ROUGE) scores.
Unlike the above method, as Fig.~\ref{pipeline-ATS} (a) shows, 
Tsvigun et al.~\cite{TsvigunLSLDKBSP22} propose an alternative query strategy for ATS based on diversity principles. This strategy, known as in-domain diversity sampling, involves selecting instances that are dissimilar from annotated documents, but similar to the core documents of the domain. Given limited annotation budget,  they can improve model performance and consistency scores.

\noindent {\adfhalfrightarrowhead} \textbf{\textit{Question Answering}}
involves answering questions about images or passages of text~\cite{DBLP:journals/corr/abs-1912-01119}. 
However, current models require large-scale training data to achieve high performance. DAL methods, such as Datamap~\cite{Outliers-ACL1} and hierarchical dialog policies~\cite{DBLP:conf/aaai/PadmakumarM21}, are designed to maximize performance with minimal labeling effort. 
Specifically, in Fig.~\ref{pipeline-ATS} (b),  DataMap~\cite{Outliers-ACL1} is able to detect and eliminate outlier examples from the unlabeled set, resulting in a significant increase in model accuracy with fewer labeled examples.
Padmakumar et al.~\cite{DBLP:conf/aaai/PadmakumarM21} develop a joint policy for clarification and DAL in an interactive image retrieval task. Asking users for clarification while querying new examples improves the model performance.

\noindent {\adfhalfrightarrowhead} \textbf{\textit{Information Extraction}} refers to many NLP tasks, including named entity recognition, keyword extraction, word segmentation, etc.
Manual annotation of large-scale sequences is time consuming, expensive, and thus difficult to realize.
To address this, Brantley et al.~\cite{BrantleyDS20} design a new DAL annotation manner. 
They use a noisy heuristic labeling function to provide initial low-quality labels, train a classifier to decide whether to trust these labels,  and annotate the most uncertain samples with trustable labels. 
Their model achieves high efficiency and effectiveness on many information extraction tasks.
Similarly, Radmard et al.~\cite{RadmardFL20} focus on improving the efficiency of DAL for naming entity recognition by querying subsequences within each sentence and propagating labels to unseen identical subsequences in the dataset. 
They demonstrate that the DAL strategy requires only 20\% of the dataset to achieve the same results as training on the full dataset.
Hua et al.~\cite{Hua022} propose two model-independent acquisition strategies for identifying and understanding the structure of argumentative discourse, achieving competitive results with fewer computations
The former selects samples with the most novel words for labeling, while the latter seeks to identify more relation links by matching any of the 18 prominent discourse markers from a manual.

\noindent {\adfhalfrightarrowhead} \textbf{\textit{Semantic Parsing}}
aims to convert a natural language utterance to a logical form: a machine-understandable representation of its meaning~\cite{MoradshahiTCL23}. 
DAL can help reduce data requirements and improve efficiency for semantic parsing.
For example, 
Duong et al.~\cite{DuongAEPCJ18} design a simple hyperparameter selection technique for DAL to accelerate data annotation.
Experiments show that their method significantly reduces the need for data annotation and improves the model's performance on semantic parsing.
Li et al.~\cite{DBLP:conf/eacl/LiH23} also design a hyperparameter tuning module to reduce the additional annotation cost.
In addition, they design a novel query strategy that prioritizes examples with various logical form structures and more lexical choices, which further improve the performance for semantic parsing.
Cohen et al.~\cite{DBLP:conf/acl/LiQCTH23} propose a novel DAL method with two new annotation manners, called HAT. 
Experiments show that HAT can pick out the most semantically varied and illustrative utterances, leading to the highest possible gains in parser performance.

\subsection{\textbf{Applications in Computer Vision}}

With the remarkable success of CNNs and Vision Transformers,
a valuable insight has been gained that more labeled image datasets can promote to obtain better performance of the task.
However, as the amount of data increases, training DNNs becomes time and resource consuming. 
Additionally, even if the number of data increases, the presence of noise often leads to limited performance improvement. 
DAL can effectively reduce noise and time consumption in many CV tasks.
Hereafter, we provide detailed information on specific tasks and their improvements achieved with DAL in CV.

\noindent {\adfhalfrightarrowhead} \textbf{\textit{Image Classification}} aims to accurately classify images based on the provided labels for many specific fields such as remote sensing~\cite{DBLP:journals/corr/abs-2104-07784}, medical imaging~\cite{DBLP:conf/cvpr/ZhouSZGGL17} and face recognition~\cite{DBLP:journals/pami/LinWMZZ18}.
We list the most successful DAL methods for image classification in Section~\ref{Import}, such as BCBA, DBAL and CEAL,  which can be referred to for more detailed information.

\noindent {\adfhalfrightarrowhead} \textbf{\textit{Image Captioning}} aims to automatically generate descriptive text about the content of an image.
Achieving high-quality captioning requires large-scale datasets with diverse images. 
Unfortunately, creating such a dataset is time-consuming and costly.
To tackle this issue, 
Zhang et al.~\cite{DBLP:conf/mm/ZhangL0WDZH20} devise a novel adversarial DAL model, which uses visual and textual information to select the most representative samples to optimize the performance of image captioning. Experiments show that they overcome the limitations of labeled data scarcity and improve the practicality and effectiveness of image captioning.
In a similar vein, Cheikh et al.~\cite{DBLP:conf/lion/CheikhZ20} introduce a knowledge-transferable DAL framework for low-resorce datasets. 
They take advantage of existing datasets, translating their captions into Arabic, and train the model with translated caption datasets as prior knowledge for low-resource ArabicFlickr1K datasets (which contain only 1,095 images). 
Their model achieves the Bilingual Evaluation Understudy (BLEU) score of 47\%, serving as compelling evidence for the effectiveness of their approach.

\noindent {\adfhalfrightarrowhead} \textbf{\textit{Semantic Segmentation}} aims to understand images at the pixel level, serving as the basis for various applications, including autonomous driving~\cite{DBLP:conf/cvpr/XieY0LC22} and
robot manipulation~\cite{ayub2022fewshot}. 
However, training segmentation models requires an extensive amount of data with pixel-wise annotations, a process that is burdensome and prohibitively expensive~\cite{DBLP:conf/iccv/WuLHLSHH21}.
To solve this challenge, 
Konyushkova et al.~\cite{DBLP:journals/cviu/KonyushkovaSF19} propose an uncertainty-based DAL method with geometric priors to expedite and simplify the annotation process for image segmentation. Experiments show that their method can be applied to both background-foreground and multi-class segmentation tasks.
Qiao et al.~\cite{DBLP:conf/aaai/QiaoZLZWDY22} introduce a collaborative panoptic regional DAL framework for partial annotated semantic segmentation. 
By incorporating semantic-agnostic panoptic matching and region-based selection and extension, their model strikes a balance between labeling efforts and performance.
Similarly, Xie et al.~\cite{DBLP:conf/cvpr/XieY0LC22} propose an automated region-based DAL approach for semantic segmentation considering the spatial adjacency of image regions and the confidence in prediction. 
Experiments show that they can use a small number of labeled image regions while maximizing segmentation performance.

\noindent {\adfhalfrightarrowhead} \textbf{\textit{Object Detection}}
is transformed into a region classification task by generating candidate regions of objects from the input image.
Features are typically extracted from candidate object regions using CNNs and classifiers are subsequently employed for the final detection. 
DAL can reduce labeled data to better fit numerous parameters of CNN.
Wu et al.~\cite{DBLP:conf/cvpr/WuC022} propose a novel hybrid query strategy that jointly considers uncertainty and diversity. Extensive experiments are conducted on two object detection datasets that effectively demonstrate the superiority and effectiveness of their model.
Wang et al.~\cite{DBLP:journals/tnn/WangLYCZZ19} introduce active sample mining with switchable selection criteria to incrementally train robust object detectors using unlabeled or partially labeled samples, avoiding the influence of noisy samples and outliers. 
The effectiveness of the model is demonstrated through extensive experiments on publicly available object detection benchmarks.
Yuan et al.~\cite{DBLP:conf/cvpr/YuanWFLXJY21} define an instance uncertainty learning module that takes advantage of the discrepancy of two adversarial instance classifiers trained in the labeled set to predict the instance uncertainty of the unlabeled set. 
With iterative instance uncertainty learning and re-weighting, they suppress noisy instances, bridging the gap between instance and image-level uncertainty.

\noindent {\adfhalfrightarrowhead} \textbf{\textit{Pose Estimation}} aims to localize the positions of specific key points in images, which has a wide range of applications, such as augmented reality, translation of sign language, and human-robot interaction. 
Obtaining pose annotations can be extremely expensive and laborious. 
To address this issue, Caramalau et al.\cite{DBLP:conf/wacv/CaramalauBK21} propose distribution-based methods for the selection of diverse and representative samples.
Experiments demonstrate their high efficiency and effectiveness for pose estimation. 
Similarly, Shukla et al.\cite{DBLP:conf/cvpr/ShuklaA21} use an
uncertainty-based query strategy and annotate samples with the lowest confidence scores and further improve the performance with fewer labeled samples.
Gong et al.~\cite{DBLP:conf/cvpr/GongFKRL22} design a novel meta agent teaming DAL (MATAL) framework to actively select and label informative images for effective learning. 
MATAL formulates the sample selection procedure as a Markov Decision Process and learns an optimal sampling policy that effectively maximizes the performance of the pose estimator.

\noindent {\adfhalfrightarrowhead} \textbf{\textit{Target Tracking}} aims to accurately track targets in images, which can be applied for numerous applications, including video surveillance, autonomous vehicles, etc. 
Using DAL can better help train neural networks with limited labeled samples for target tracking.
Yuan et al.~\cite{DBLP:journals/corr/abs-2110-13259} present a new DAL sequence selection method in a multi-frame collaboration way for target tracking. 
To ensure the diversity of selected sequences, they measure samples' similarity by their temporal relation between multiple frames in each video, and they use a nearest neighbor discriminator to select the representative samples. 
Experiments show that their method can eliminate background noise and improve efficiency.

\noindent {\adfhalfrightarrowhead} \textbf{\textit{Person Re-identification}} (Re-ID)
aims to match a specific pedestrian using different cameras, which is an essential task for public security. 
Previous efforts mainly concentrate on enhancing the performance of Re-ID models, relying on large labeled datasets. However, these efforts often overlook data redundancy issues that can arise in constructing Re-ID datasets.
To address data redundancy in Re-ID datasets, 
Liu et al.~\cite{DBLP:conf/iccv/LiuWGTL19} propose an alternative human-in-the-loop model based on reinforce learning. In their method, a human annotator provides binary feedback to fine-tune a pre-trained CNNs Re-ID model. Extensive experiments prove the superiority of their method compared to existing unsupervised, transfer learning, and DAL models.
On the other hand, Xu et al.~\cite{DBLP:journals/pr/XuLZGH21} focus on learning from scratch with incremental labeling through human annotators and model feedback. They combine DAL with an incremental annotation process to select informative and diverse samples without redundancy from an unlabeled set in each iteration. These samples are then labeled by human annotators to further improve the performance of the model.

\begin{figure}[h]
\includegraphics[width=0.48\textwidth]{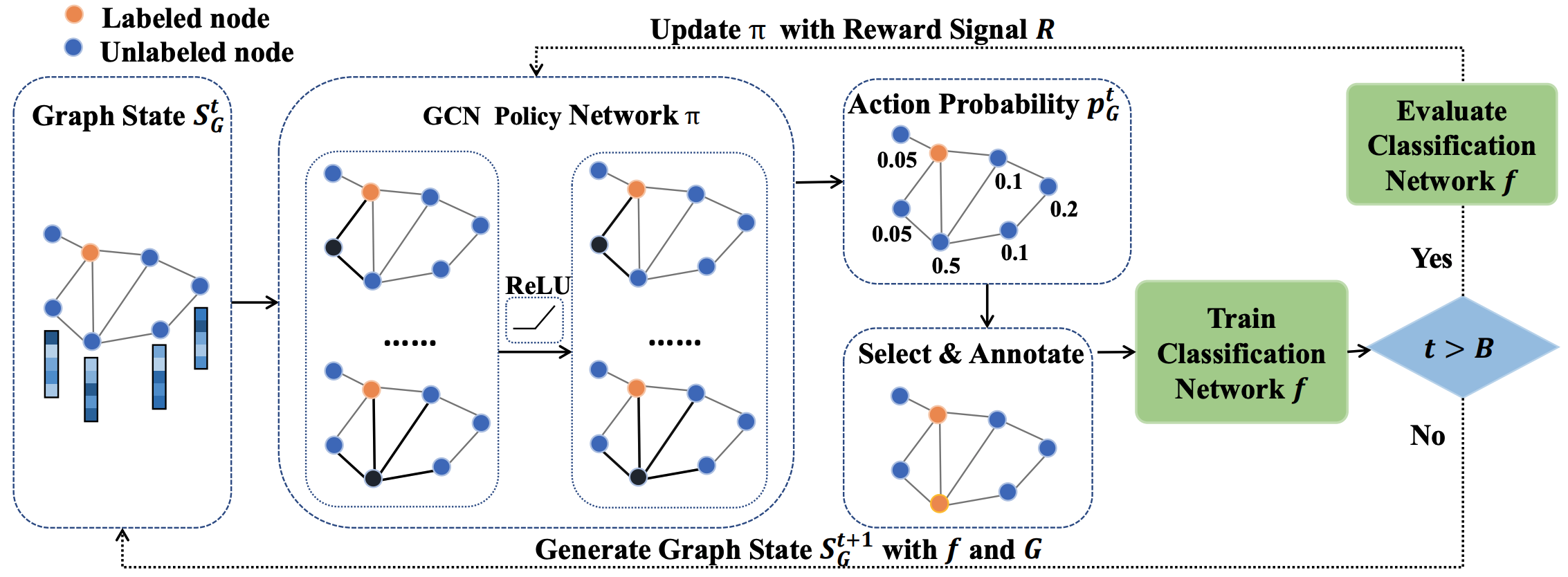}
\caption{The framework of Graph Policy {Network}~\cite{DBLP:conf/nips/HuXQYC0T20}.}
\label{pipeline-he}
\end{figure}

\subsection{{\textbf{Applications in Graph Data Mining and Learning}}}
There are substantial increase in content-rich network from various domains, such as social networks, citation networks, and financial networks. 
Graphs have emerged as a powerful tool for representing and discovering knowledge, with nodes representing instances characterized by rich content features and edges denoting relationships or interactions between nodes.

\noindent {\adfhalfrightarrowhead} \textbf{\textit{Node Classification}} is to predict the labels of unlabeled nodes in a partially labeled network. 
GNNs rely heavily on a sufficient number of labeled nodes, which is costly and time-consuming.
To address this problem, many graph-based DAL methods are proposed.
For example, 
ICA-based methods~\cite{DBLP:conf/icml/BilgicMG10} leverage label dependence among neighboring nodes to select diverse samples for node classification, while AGE~\cite{DBLP:journals/corr/CaiZC17} and ANRMAB~\cite{DBLP:conf/ijcai/GaoYZWPH18} integrate GCNs with three traditional DAL query strategies and achieve good performance on many node classification datasets.
%
As Fig.~\ref{pipeline-he} shows, Hu et al.~\cite{DBLP:conf/nips/HuXQYC0T20} present a graph policy network for transferable DAL on graphs, which formalizes DAL on graphs as a Markov decision process and learns the optimal query strategy with reinforce learning. The state is defined based on the current graph status, and the action is to select a node for annotation at each query step. The reward is defined as the performance gain of the GNNs trained with the selected nodes.

\noindent {\adfhalfrightarrowhead} \textbf{\textit{Link Prediction}} aims to predict missing or potential links between nodes in a given network. It involves using existing connections or relationships to infer the likelihood of forming new connections. 
In the context of link prediction, the challenge arises from the limited availability of existing link information between nodes in a network. 
DAL can help alleviate this issue, for example,
DALAUP~\cite{DBLP:conf/ijcai/ChengZY0LT019} uses neural networks to obtain vector representations of user pairs and utilizes multiple query strategies to select informative user pairs for labeling and model training, achieving superior performance compared to existing methods. 
Cai et al.\cite{DBLP:conf/ijcai/CaiTCHWH19} design a multi-view DAL method that reduces the annotation cost by selectively querying metadata for the most informative examples, using a mapping function from the visual view to the text view. 
They demonstrate that multi-view DAL can use richer information to help improve performance than using single view.
Zhao et al.\cite{DBLP:journals/ai/ZhaoPY17} propose a DAL-based transfer learning framework for link prediction in recommender systems, which iteratively selects entities from source systems for target systems using uncertainty-based criteria.
Experiments show that their method successfully improves efficiency and effectiveness.

\noindent {\adfhalfrightarrowhead} \textbf{\textit{Community Detection}} 
aims to accurately partition nodes into distinct classes based on the topological structure of the networks.
However, in many practical scenarios, unsupervised methods struggle to achieve the exact community.
To solve this issue,
Gupta et al.~\cite{DBLP:journals/grid/GuptaJS18} propose community trolling, a DAL-based method for topic-based community detection. Their method selects relevant samples from polluted big data, reducing the unreliable dataset to a reliable one for studying communities.
%
Chien et al.~\cite{DBLP:conf/aaai/ChienTL20} propose a novel DAL method for geometric community detection.
They first remove many cross-cluster edges while preserving intra-cluster connectivity to avoid noise.
Then, they interactively query the label of one node for each disjoint component to recover the underlying clusters.
Experiments show that they can achieve SOTA performance in community detection.

\subsection{\textbf{Other Selected Interesting Applications}}

\noindent  {\adfhalfrightarrowhead} \textbf{\textit{Engineering Systems.}} 
DAL methods exhibit remarkable performance in computationally demanding engineering systems by significantly reducing running time and computational costs. For example, Yue et al.~\cite{DBLP:journals/tase/YueWHS21} introduce two novel DAL algorithms: the variance-based weighted AL and the D-optimal weighted AL, designed specifically for Gaussian processes with uncertainties. Numerical studies demonstrate the effectiveness of their approach, notably improving predictive modeling for automatic shape control of composite fuselage structures.
In another vein, Lee et al.~\cite{DBLP:journals/tase/LeeWWY23} optimize their DAL acquisition function by jointly considering safe variance reduction and safe region expansion tasks, aiming to minimize failures without explicit knowledge of failure regions. This approach is tailored for real systems with uncertain failure conditions, as demonstrated in the predictive modeling of composite fuselage deformation, achieving zero failures by considering the composite failure criterion.
Furthermore, Lee et al.~\cite{DBLP:journals/jcise/LeeWWCY23} introduce a partitioned DAL method, comprising two systematic steps: global searching for uncertain design spaces and local searching using local Gaussian processes. They apply their method to aerospace manufacturing and materials science, achieving superior performance in prediction accuracy and computational efficiency compared to benchmarks.

\noindent  {\adfhalfrightarrowhead} \textbf{\textit{{Personalized Medical Treatment}}} explores how patient health is affected by taking a drug and how user questions are answered by search recommendation~\cite{DBLP:journals/nms/Rahman20}. 
Although modern methods can achieve impressive performance, they need a significant amount of labeled data. 
To solve this issue,
Deng et al.~\cite{DBLP:conf/adprl/DengPM11} propose the use of DAL to recruit patients and assign treatments that reduce the uncertainty of an Individual Treatment Effect model.
Sundin et al.~\cite{DBLP:conf/icml/SundinSSVSK19} propose to use a Gaussian process to model the individual treatment effect and use the expected information gain over the S-type error rate, defined as the error in predicting the sign of the conditional average treatment effect, as their acquisition function.
Jesson et al.~\cite{DBLP:conf/nips/JessonTAKSG21} develop epistemic uncertainty-aware methods for DAL of personalized treatment effects from high-dimensional observational data. In contrast to previous work that only uses information gain as the acquisition objective, they propose Causal-BALD  because they consider both information gain and overlap between the treatment and control groups. 
Li et al.~\cite{10389652} used DAL to help people by recognizing their emotion.

\begin{table*}[]
\centering
\caption{Summary of various challenges and opportunities.}
\begin{tabular}{@{}lll@{}}
\toprule
\multicolumn{1}{c}{\textbf{Challenge Types}}                            & \multicolumn{1}{c}{\textbf{Challenges}}                               & \multicolumn{1}{c}{\textbf{Opportunities}}       \\ 
\midrule
\multirow{10}{*}{\textbf{Pipeline-related Issues}} & 
\multicolumn{1}{l}{\multirow{3}{*}{Inefficient \& Costly human annotation}}       & servers, workers and annotators share information~\cite{DBLP:conf/ijcai/HuangZNY21}.  \\
\multicolumn{1}{c}{}  & \multicolumn{1}{l}{}    & self-supervised pseudo-labels to reduce human efforts~\cite{DBLP:journals/tcsv/WangZLZL17, DBLP:journals/pr/YangL19a
}.         \\
\multicolumn{1}{c}{}  & \multicolumn{1}{l}{}    & incorporate additional knowledge to reduce expert knowledge~\cite{DBLP:journals/pr/YangL22,DBLP:conf/aaai/QiaoZLZWDY22
}.      \\
\specialrule{0em}{2pt}{2pt}
\cmidrule(l){2-3}
\multicolumn{1}{c}{}  & \multirow{3}{*}{Insufficient research on stopping strategies} & the confidence among the selected samples does not increase~\cite{DBLP:conf/sigir/McDonaldMO20}.     \\
\multicolumn{1}{c}{}  &   & stop when all instances lie between two contour lines~\cite{DBLP:journals/tnn/YuYZS19}.   \\
\multicolumn{1}{c}{}  &   & upper bound in expected generalization errors as stopping criterion~\cite{DBLP:conf/aistats/IshibashiH20}. \\
\specialrule{0em}{2pt}{2pt}
\cmidrule(l){2-3}
\multicolumn{1}{c}{}  & \multirow{3}{*}{Cold-start}              & use pre-trained embeddings~\cite{DBLP:conf/emnlp/YuanLB20,DBLP:conf/acl/YuZXZSZ23}.                \\
\multicolumn{1}{c}{}  &   & design initial queries~\cite{DBLP:journals/corr/abs-2210-02442,yehuda2022active}.               \\
\multicolumn{1}{c}{}  &   & use diverse sampling~\cite{DBLP:journals/tcyb/CaoTX22,DBLP:conf/iclr/MahmoodFL22}. \\
\midrule
\multirow{10}{*}{\textbf{Tasks-related Issues}}   & \multirow{3}{*}{Difficulty in cross-domain transfer}          & select samples in regions of joint disagreement between models~\cite{DBLP:conf/aaai/ZhouC0GY21,tang2022active,DBLP:journals/corr/abs-2205-03555}.               \\
 &   & source and target domain distribution matching~\cite{vu-etal-2019-learning,farquhar2021on}.         \\
 &   & transferable DAL policies between the source and target graphs~\cite{DBLP:conf/nips/HuXQYC0T20}.    \\
\specialrule{0em}{2pt}{2pt}
\cmidrule(l){2-3}
 & \multirow{3}{*}{Unstable performance}    & avoid DAL's sensitivity to the initial labeled set~\cite{DBLP:journals/tnn/YuYZS19,DBLP:conf/sigir/Zlabinger19,DBLP:journals/pr/YangL22,DBLP:conf/emnlp/YuanLB20,DBLP:conf/emnlp/Ein-DorHGSDCDAK20}.                 \\
 &   & use distribution information to improve model's robustness~\cite{DBLP:conf/aaai/KwakKKHY22,DBLP:journals/corr/abs-2111-04286,DBLP:journals/corr/abs-2111-04286}.              \\
 &   & use pre-trained language model~\cite{DBLP:conf/acl/SchroderNP22,mamooler-etal-2022-efficient}.        \\
\specialrule{0em}{2pt}{2pt}
\cmidrule(l){2-3}
 & \multirow{3}{*}{Lack of scalability \& generalizability}      & hybrid strategies for sample selection~\cite{Maekawa,DBLP:journals/pr/ZhaoSZCG19}.             \\
 &   & nearest-neighbor classifiers~\cite{DBLP:conf/aaai/WanYFJHY21}.             \\
 &   & combining annotation and counterfactual sample construction~\cite{DBLP:conf/acl/DengWFZ0L23,DBLP:conf/nips/WangHWTMH22}.          \\
 \midrule
\multirow{10}{*}{\textbf{Datasets-related Issues}}                     & \multirow{3}{*}{Outlier Data \& Noisy Oracles}                & find the best balance between purity and informativeness~\cite{DBLP:conf/nips/ParkSBLS022,DBLP:conf/nips/ElenterNR22}.                     \\
 &   & knowledge distillation~\cite{DBLP:conf/iccv/PengWLY21}.                    \\
 &   & relabeling frameworks for correst oracle labels~\cite{DBLP:journals/tnn/YuTWDZ21,DBLP:conf/socialcom/ZhaoSS11,DBLP:conf/ijcai/AshariG19}.     \\
\specialrule{0em}{2pt}{2pt}
\cmidrule(l){2-3} 
& \multirow{3}{*}{Data Scarcity \& Imbalance}                   & data augmentation and large PLMs ~\cite{DBLP:conf/eccv/ChenZWCL22,DBLP:conf/aaai/SeoKAL22,DBLP:conf/cvpr/GudovskiyHYT20}.   \\
 &   & cost-sensitive learning~\cite{DBLP:journals/tnn/YuYZS19,DBLP:conf/cvpr/ChoiYKCKCGC21}.              \\
 &   & design new query strategies for imbalanced datasets~\cite{DBLP:conf/aistats/ZhaoLAY21,DBLP:conf/nips/HartfordLRPLL20,DBLP:conf/icml/ZhangKN22}.                     \\
\specialrule{0em}{2pt}{2pt}
\cmidrule(l){2-3}
 & \multirow{3}{*}{Class distribution mismatch}                  & new DAL query strategy~\cite{DBLP:journals/pami/DuCZCCL23,DBLP:conf/cvpr/HeHLY22}.               \\
 &   & new DAL framework~\cite{DBLP:conf/ijcai/TangH21}.    \\
 &   & incoporate additional detector~\cite{DBLP:conf/cvpr/NingZLH22}.           \\ \bottomrule
\end{tabular}
\label{table: summary of all challenges}

\end{table*}

\section{Challenges \& Opportunities of DAL}
\label{Challenge}
As Table~\ref{table: summary of all challenges} shows, hereafter, we summarize the challenges and the corresponding potential solutions and opportunities. 

\subsection{\textbf{Pipeline-related Issues}}

\noindent  {\adfhalfrightarrowhead} \textbf{\textit{{Inefficient \& Costly Human Annotation.}}}
DAL assumes that human annotators are readily available to label new samples once they are required.
However, this assumption may not hold in some real-world applications.
Human annotators can get tired or need breaks, forcing the DAL process to be suspended until they reappear. 
Moreover, human annotation is time-consuming and needs expert knowledge, resulting in long waits before models can be re-trained with new labeled data.

\begin{figure}[t]
\centering
\includegraphics[width=0.48\textwidth]{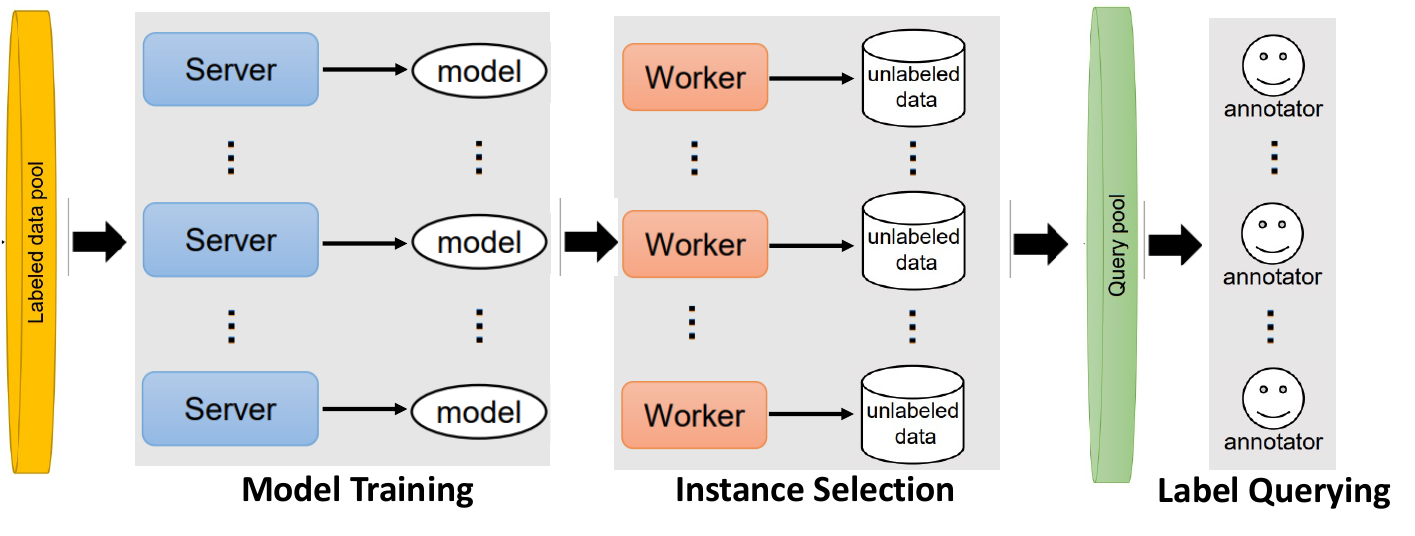}
\caption{The framework for efficiently annotation.}
\label{pipeline-An}
\end{figure}

To improve efficiency, DAL methods incorporate additional techniques to reduce human annotation. 
Wang et al.~\cite{DBLP:journals/tcsv/WangZLZL17} use self-supervised learning by adding pseudo-labels with high confidence to help reduce human effort and improve the performance of the model. 
Go one step further, Yang et al.~\cite{DBLP:journals/pr/YangL19a} introduce multiple pseudo-annotators that provide labels for unlabeled samples, achieving good performance without requiring human expert knowledge.
On the other hand, 
as shown in Fig.~\ref{pipeline-An}, Huang et al.~\cite{DBLP:conf/ijcai/HuangZNY21} propose a new annotation strategy to allow servers, workers, and annotators to cooperate efficiently for sharing candidate queries and annotations. Experiments show that their model can avoid annotation noise and save much time for re-checking annotations.
To further reduce expert knowledge,
others tend to reduce the search scope in each iteration to improve efficiency. 
For example, 
Yang et al.~\cite{DBLP:journals/pr/YangL22} restrict candidate samples to their nearest neighbors of the labeled set rather than scanning all data. 

\noindent  {\adfhalfrightarrowhead} \textbf{\textit{{Insufficient Research on Stopping Strategies.}}}
Few studies are designed for stopping strategies of DAL methods~\cite{DBLP:conf/nips/HartfordLRPLL20}. 
However, stopping strategies are essential for DAL because they reduce the amount of human labor by limiting the number of samples that need to be labeled and prevent the inclusion of noisy and redundant samples, which can negatively affect the performance of DAL models.

McDonald et al.~\cite{DBLP:conf/sigir/McDonaldMO20} design two novel stopping strategies for DAL methods in the document classification task. 
The first strategy measures the overall confidence of the classifiers in correctly classifying the remaining unlabeled documents. 
It assumes that when the classifier's mean confidence level for the remaining documents stabilizes, the model stops the DAL process, since its effectiveness would no longer improve.
The second strategy measures the confidence of the classifiers among the selected documents to be reviewed. It assumes that when the classifier's confidence stops increasing for these documents, it has reached its maximal confidence and stops the DAL process.
Benefiting from the idea of the margin exhaustion criterion, Yu et al.~\cite{DBLP:journals/tnn/YuYZS19} identify two corresponding contour lines in the instance space and assume that the DAL process can only be stopped when all instances lying between these two contour lines have been labeled.
They achieve good performance in many classification tasks.
Based on the Bayesian theory,  Ishibashi et al.~\cite{DBLP:conf/aistats/IshibashiH20} derive a novel upper bound for the difference in expected generalization errors before and after obtaining new training data. They then combine this upper bound with a statistical test to derive a stopping criterion for DAL and significantly improve efficiency.

\noindent  {\adfhalfrightarrowhead} \textbf{\textit{{Cold-start.}}}
Most DAL methods fail to improve over random selection when the annotation budget is very small, a phenomenon sometimes term as ``cold-start''~\cite{DBLP:journals/corr/abs-2210-02442}. 
Uncertainty sampling has been shown to be inherently unsuitable for low budgets, possibly explaining the cold-start phenomenon~\cite{DBLP:conf/icml/HacohenDW22}. 
Low budgets can be seen in many applications, especially those that require an expert tagger whose time is expensive. 
If we want to expand deep learning to new domains, overcoming the cold-start problem is an ever-important task.

\begin{figure}[h]
\centering
\includegraphics[width=0.48\textwidth]{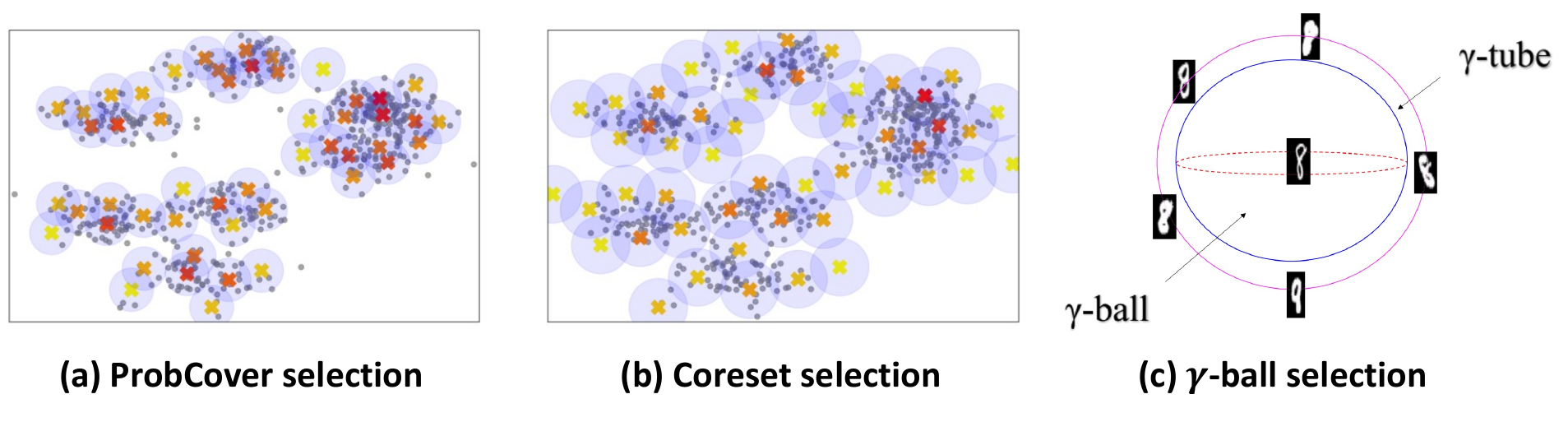}
\caption{An example for cold-start data selection.}
\label{pipeline-cold}
\end{figure}

To relieve the cold-start issue, 
Yuan et al.~\cite{DBLP:conf/emnlp/YuanLB20} use pre-trained embeddings on unsupervised tasks, decreasing budget dependency while remaining faithful to uncertainty sampling.
Similarly, Yu et al.~\cite{DBLP:conf/acl/YuZXZSZ23} try to use pre-trained knowledge from PLMs to avoid cold-start.
They select few shot samples to fine-tune large-scale PLM,  achieve SOTA performance in six datasets, and improve the efficiency of labeling over existing baselines by 3.2\%–6.9\% on average.
On the other hand, 
in Fig.~\ref{pipeline-cold} (a-b), Yehuda et al.~\cite{yehuda2022active} develop a new DAL initialization strategy to solve the cold-start issue for low-budget image classification, which significantly outperforms CoreSet initialization in the low-budget regime. 
They also theoretically analyze different DAL strategies in embedding spaces and improve performance on both low- and high-budget scenes.
In Fig.~\ref{pipeline-cold} (c), Cao et al.~\cite{DBLP:journals/tcyb/CaoTX22}  apply the informative sampling policy on the $\gamma$ tube to solve the cold-start sampling problem.
Mahmood et al.~\cite{DBLP:conf/iclr/MahmoodFL22} query a diverse set of examples with minimal Wasserstein distance from unlabeled data. They report a significant performance boost in the low-budget regime.

\subsection{\textbf{Task-related Issues}}\label{task-related}

\noindent  {\adfhalfrightarrowhead} \textbf{\textit{{Difficulty in Cross-domain Transfer.}}}
We discuss two difficulties of cross-domain transfer in DAL. 
First, machine learning systems are always deployed on various devices with the same labeled dataset. 
However, DAL is often model-dependent and not directly transferable, i.e., data queried for one model may be less effective for another~\cite{tang2022active}; 
Second, transfer learning biases DAL to select samples that match the distribution of the source domain to the target domain, leading to sampling bias and the high cost of transfer learning.

\begin{figure}[h]
\includegraphics[width=0.48\textwidth]{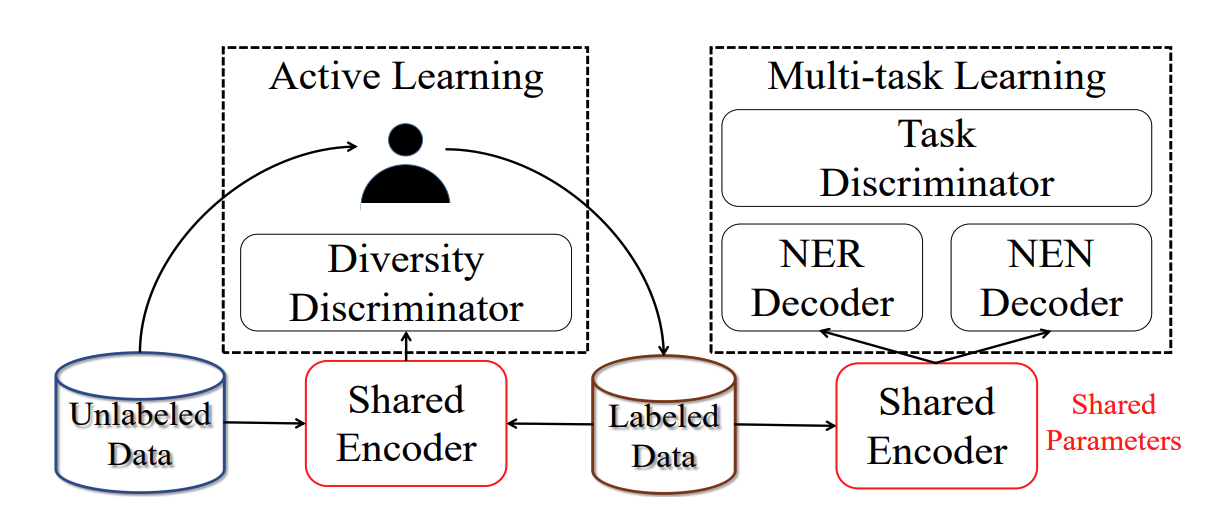}
\caption{Multi-task learning transfer knowledge from sources~\cite{DBLP:conf/aaai/ZhouC0GY21}.}
\label{pipeline-Mul}
\end{figure}

To benefit multiple target models, 
some methods aim to select samples in joint disagreement regions across models~\cite{tang2022active}, adopt multi-agent reinforcement learning for optimal selection~\cite{DBLP:journals/corr/abs-2205-03555}, or leverage multi-task learning to transfer common knowledge from the source domain as shown in Fig.~\ref{pipeline-Mul}.
To avoid sampling bias, 
Farquhar et al.~\cite{farquhar2021on} apply corrective weighting using an unbiased risk estimator to maintain the target distribution during pool-based sampling.
Trang et al.~\cite{vu-etal-2019-learning} introduce a heuristic query strategy that matches the distribution of the source domain while retrieving valuable target samples.
Hu et al.~\cite{DBLP:conf/nips/HuXQYC0T20} learn transferable DAL policies on labeled source graphs that generalize selection to unlabeled target graphs.
Experiments show that the above methods can achieve excellent performance and transferability.

\noindent  {\adfhalfrightarrowhead} \textbf{\textit{{Unstable Performance.}}} 
DAL methods always have unstable performance, i.e., results for the same method vary significantly with different initialized seeds~\cite{DBLP:conf/cvpr/YooK19}.
Two primary reasons can explain this instability.
First, the DAL methods are sensitive to the initial labeled dataset.
The initial selected samples have a great influence on the eventual outcome of the current approaches. With insufficient initial labeling, subsequent DAL cycles become highly biased, resulting in poor selection.  
Second, current DAL methods always separate active learning and deep learning methods into two separate processes, easily leading to sub-optimal and unstable performance~\cite{mosbach2021on}.

To solve DAL's sensitivity to the initialization, current methods always use diverse sampling and pre-trained models.
Yu et al.~\cite{DBLP:journals/tnn/YuYZS19} adopt hierarchical clustering to select 10\% samples near each clustering center as representative samples. Their new initialization greatly helps stabilize the performance. 
Zlabinger et al.~\cite{DBLP:conf/sigir/Zlabinger19} take into account both diversity and polarization to effectively select initial samples for DAL methods that further stabilize the performance of the DAL process.
Yang et al.~\cite{DBLP:journals/pr/YangL22} select initial samples by evaluating the total distance between the unlabeled samples and the initial samples, showing that the same distance between them can result in better and stable performance.
On the other hand, Yuan et al.~\cite{DBLP:conf/emnlp/YuanLB20} incorporate language information as prior knowledge to help learn node representations and use clustering methods to select the initial data.
Similarly, Ein-Dor~\cite{DBLP:conf/emnlp/Ein-DorHGSDCDAK20} uses BERT to learn the representations of the input sentences and uses a hybrid query strategy to select the most uncertain and diverse samples as the initialized training data.

\begin{figure}[h]
\centering
\includegraphics[width=0.48\textwidth]{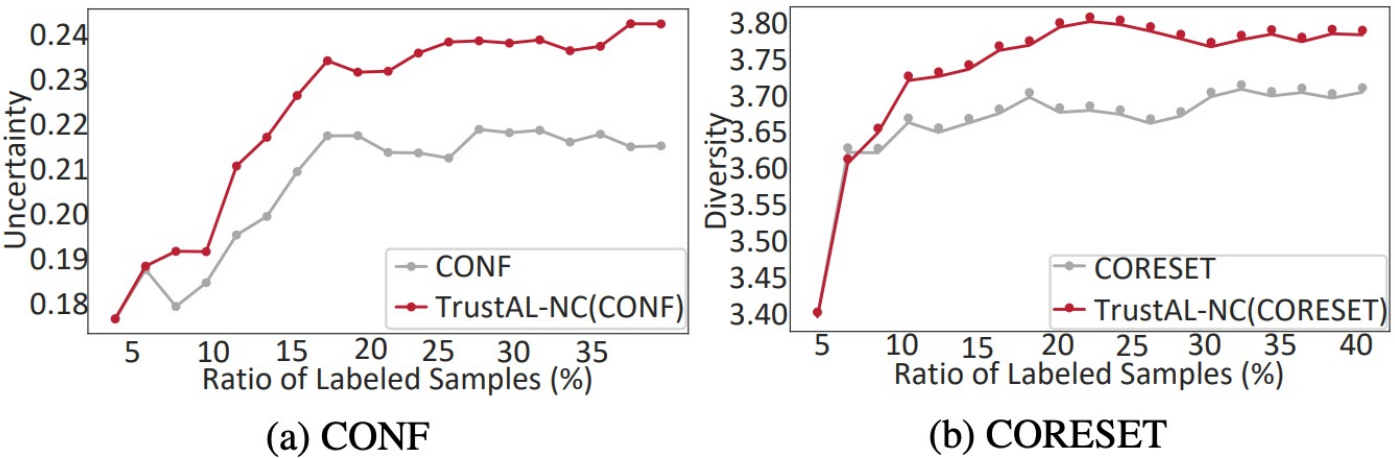}
\caption{Stable performance of TrustAL~\cite{DBLP:conf/aaai/KwakKKHY22}.}
\label{pipeline-TrustAL}
\end{figure}

To bridge the gap between AL and deep learning models,
Kwak et al.~\cite{DBLP:conf/aaai/KwakKKHY22} introduce Trustworthy AL (TrustAL), a label-efficient DAL framework by transferring distilled knowledge from deep learning models to the data selection process.
As Fig.~\ref{pipeline-TrustAL} shows, they jointly optimize knowledge distillation and DAL to obtain a more consistent and reliable performance compared to the two best performing baselines on three benchmarks.
Similarly, 
Ma et al.~\cite{DBLP:journals/corr/abs-2111-04286} learn nonlinear embeddings to map inputs into a latent space and introduce a selection block to choose representative samples in the learned latent space to achieve stable performance.
Margatina et al.~\cite{DBLP:conf/acl/SchroderNP22} extend the PLMs to continually pre-train on available unlabeled data to tailor it to the task-specific domain, where they can benefit from both labeled and unlabeled data at each DAL iteration. Their experiments show considerable enhancements in data efficiency and stability compared to the standard fine-tuning approach, emphasizing the importance of a suitable training strategy in DAL.
Mamooler et al.~\cite{mamooler-etal-2022-efficient} try to combine DAL with PLMs in the legal domain, where they use unlabeled data in three stages: training the model to adjust it to the downstream task, using knowledge distillation to direct the embeddings to a semantically meaningful space, and identifying the initial set.

\noindent  {\adfhalfrightarrowhead} \textbf{\textit{{Lack of Scalability \& Generalizability.}}}
Current DAL methods lack scalability, as they always require significant modifications to neural network architectures for adapting to different query strategies.
Another issue with current methods is their heavy reliance on DAL's weight parameters, while the parameters may not be generalizable to different datasets. 
Users are required to prepare additional labeled samples as a validation set to tune parameters by cross-validation, which contradicts the goal of minimizing the need for labeled data.

\begin{figure}[t]
\centering
\includegraphics[width=0.48\textwidth]{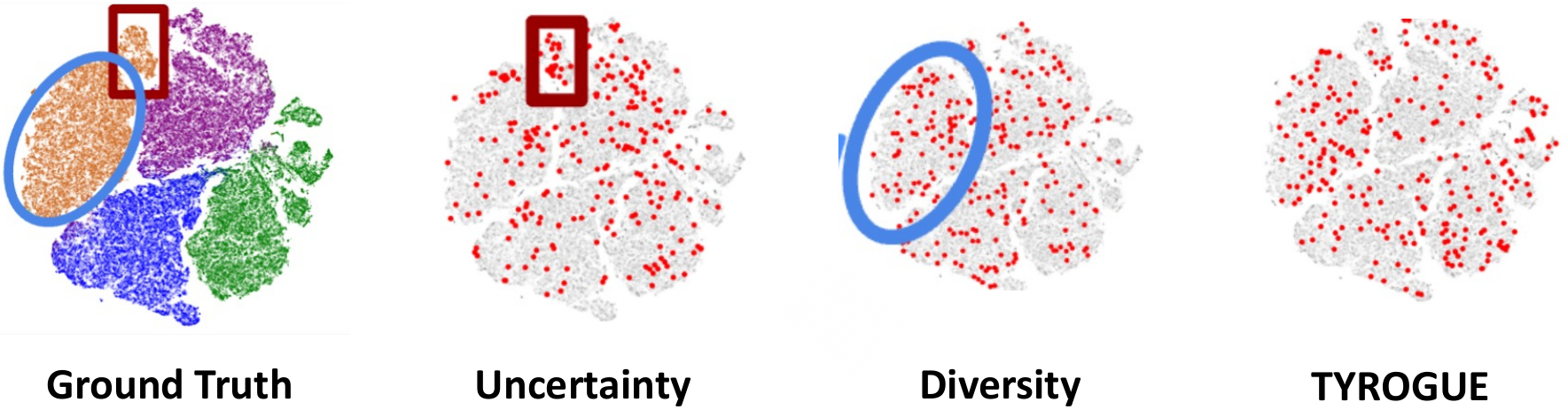}
\caption{TYROGUE can select better samples than baselines.}
\label{pipeline-tyro}
\end{figure}

In response to the above issues,
Maekawa et al.~\cite{Maekawa} introduce a novel DAL method, called TYROGUE, that uses a hybrid query strategy to improve model generalization and reduce labeling costs. 
As Figure~\ref{pipeline-tyro} shows,  uncertainty-based methods tend to acquire similar data points from a specific area within an iteration, diversity-based
methods tend to acquire data points similar to the samples acquired in previous iterations, and TYROGUE balances diversity and uncertainty by acquiring samples that are diverse and
also closer to the model decision boundary.
RMQCAL~\cite{DBLP:journals/pr/ZhaoSZCG19} is a novel scalable DAL method, which allows for any number and type of query criteria, eliminates the need for empirical parameters, and makes the trade-offs between the query criteria self-adaptive. 
On the other hand,
Wan et al.~\cite{DBLP:conf/aaai/WanYFJHY21} propose an embedded network of nearest-neighbor classifiers to enhance the generalization ability of models trained in labeled and unlabeled sub-spaces in a simple but effective manner.
Deng et al.~\cite{DBLP:conf/acl/DengWFZ0L23} focus on combining sample annotation and counterfactual sample construction in the DAL procedure to enhance the model's out-of-distribution generalization.
Wang et al.~\cite{DBLP:conf/nips/WangHWTMH22} introduce a new training manner to improve model's generalizability and show a strong positive correlation between convergence speed and generalization performance under ultra-wide conditions.

\subsection{\textbf{Dataset-related Issues}}\label{data-related}

\noindent  {\adfhalfrightarrowhead} \textbf{\textit{{Outlier Data \& Noisy Oracles.}}}
DAL methods tend to acquire outliers since models always assign high uncertainty scores to outliers.
Outliers can damage a model's learning ability and fuel a vicious cycle in which DAL methods continue to select them~\cite{DBLP:journals/tnn/WangLYCZZ19}.
Identifying and removing outliers has become an important direction in improving DAL performance and robustness.
On the other hand, 
classic DAL methods assume that annotators have high labeling accuracy.
However, in real-world settings, sample difficulty and annotator expertise can significantly affect the quality and accuracy of annotation, which may further degrade model performance.

To remove outliers, 
Park et al.~\cite{DBLP:conf/nips/ParkSBLS022} propose MQ-Net to
adaptively find the best balance between purity and informativeness of samples, filtering out noisy open-set data.
Elenter et al.~\cite{DBLP:conf/nips/ElenterNR22} introduce a new query strategy based on Lagrangian duality to select diverse samples,  efficiently removing redundant data.
Other studies~\cite{DBLP:conf/iccv/PengWLY21}
use knowledge distillation to compress useful knowledge into a small model, effectively identifying and removing outliers.
To make high-quality annotations, 
AMCC~\cite{DBLP:journals/tnn/YuTWDZ21} measures worker annotations considering both their commonality and individuality to reduce the impact of unreliable workers and improve effectiveness.
Zhao et al.~\cite{DBLP:conf/socialcom/ZhaoSS11} actively select samples that are relabeled multiple times through crowd-sourcing majority voting.
EMMA~\cite{DBLP:conf/ijcai/AshariG19} relabels samples to remove noisy annotations by analyzing the stimulus based on model memory retention and greedy heuristics.
BALT~\cite{DBLP:conf/ijcai/Tang19} improves human expertise during labeling to improve relabel quality and significantly improve model performance.
Zlabinger~\cite{DBLP:conf/sigir/Zlabinger19} trains human annotators on a set of pre-labeled samples to improve the quality of annotations. 
Huang et al.~\cite{DBLP:conf/ijcai/HuangZNY21} propose a multi-server, multi-worker framework for DAL, where
servers and workers cooperate to select diverse samples and improve model performance.

\noindent  {\adfhalfrightarrowhead} \textbf{\textit{{Data Scarcity \& Imbalance.}}}
Data scarcity poses two critical challenges.
First, datasets are difficult to collect and annotate~\cite{DBLP:conf/nips/KothawadeBKI21};
Second, DAL methods have the common underlying assumption that all classes are equal, while some classes have more samples than others (skewed class distribution~\cite{DBLP:journals/tnn/YuYZS19}) or some classes may be more difficult to learn than others, leading to sampling bias in the acquisition process~\cite{DBLP:conf/sigir/ErtekinHG07}.

\begin{figure}[t]
\centering
\includegraphics[width=0.47\textwidth]{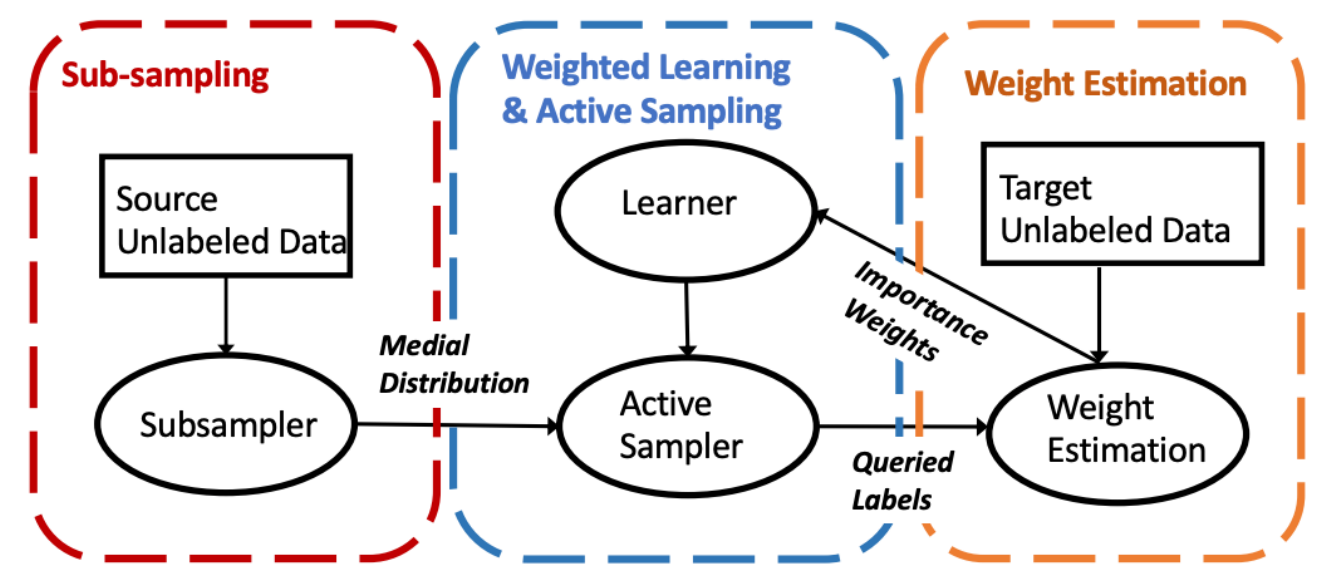}
\caption{An example of imbalanced sampling~\cite{DBLP:conf/aistats/ZhaoLAY21}.}
\label{pipeline-imbalance}
\end{figure}

For scarce datasets, 
Chen et al.~\cite{DBLP:conf/eccv/ChenZWCL22} used data augmentation to generate diverse samples to expand training data. 
Other studies used PLMs as prior knowledge and fine-tuned them to reduce the
required labeled samples~\cite{DBLP:conf/aaai/SeoKAL22}. 
For difficult annotations, 
Gudovskiy et al.~\cite{DBLP:conf/cvpr/GudovskiyHYT20} introduce several novel self-supervised pseudo-labels estimators to correct acquisition bias by minimizing the distribution shift between unlabeled data and weakly labeled validation data.
To mitigate the classes imbalance, 
Yu et al.~\cite{DBLP:journals/tnn/YuYZS19} are the first to use cost-sensitive learning. They choose the extreme weighted learning machine as the base learner to select samples based on the class imbalance ratio, class overlap, and small disjunction.
They investigate why DAL can be impacted by a skewed instance distribution and improve DAL performance on imbalanced datasets.
Choi et al.~\cite{DBLP:conf/cvpr/ChoiYKCKCGC21} solve the issue of data imbalance by considering
the probability of mislabeling a class, the probability of the data given a predicted class, and the prior probability of the abundance of a predicted class, during querying samples of DAL.
Experiments show that they can significantly enhance the ability of existing DAL methods to handle unbalanced datasets.
As shown in Fig.~\ref{pipeline-imbalance},
Zhao et al.~\cite{DBLP:conf/aistats/ZhaoLAY21} propose an alternate query strategy by using the medial distribution to find a compromise between importance weighting and class-balanced sampling. 
Experiments show that their model can be easily combined with various DAL methods and successfully select balanced samples in imbalanced datasets.
Hartford et al.~\cite{DBLP:conf/nips/HartfordLRPLL20} present an exemplar guided DAL method that shows strong empirical performance under extremely skewed label distributions by using exemplar embedding.
Zhang et al.~\cite{DBLP:conf/icml/ZhangKN22} propose a graph-based DAL method that applies a more sophisticated version of uncertainty sampling. 
Their strategy can select more evenly distributed examples for labeling than standard uncertainty sampling.

\noindent  {\adfhalfrightarrowhead} \textbf{\textit{{Class Distribution Mismatch.}}}
DAL methods assume that the labeled and unlabeled data are drawn from the same class distribution, which means that the categories of both datasets are identical~\cite{DBLP:conf/cvpr/NingZLH22}. 
However, in real-world scenarios, unlabeled data often come from uncontrolled sources, and a large portion of the examples may belong to unknown classes. 
For example,
when crawling images for binary image classification using keywords like ``dog'' and ``cat,'' over 50\% of the images in the unlabeled dataset are irrelevant to the task (e.g., ``deer,'' ``horse''). 
Annotating these irrelevant images will lead to a waste of annotation budget as they are unnecessary for training the desired classifier. Despite this challenge, existing DAL systems tend to select these irrelevant images for annotation, as they contain more uncertain knowledge.

\begin{figure}[t]
\centering
\includegraphics[width=0.48\textwidth]{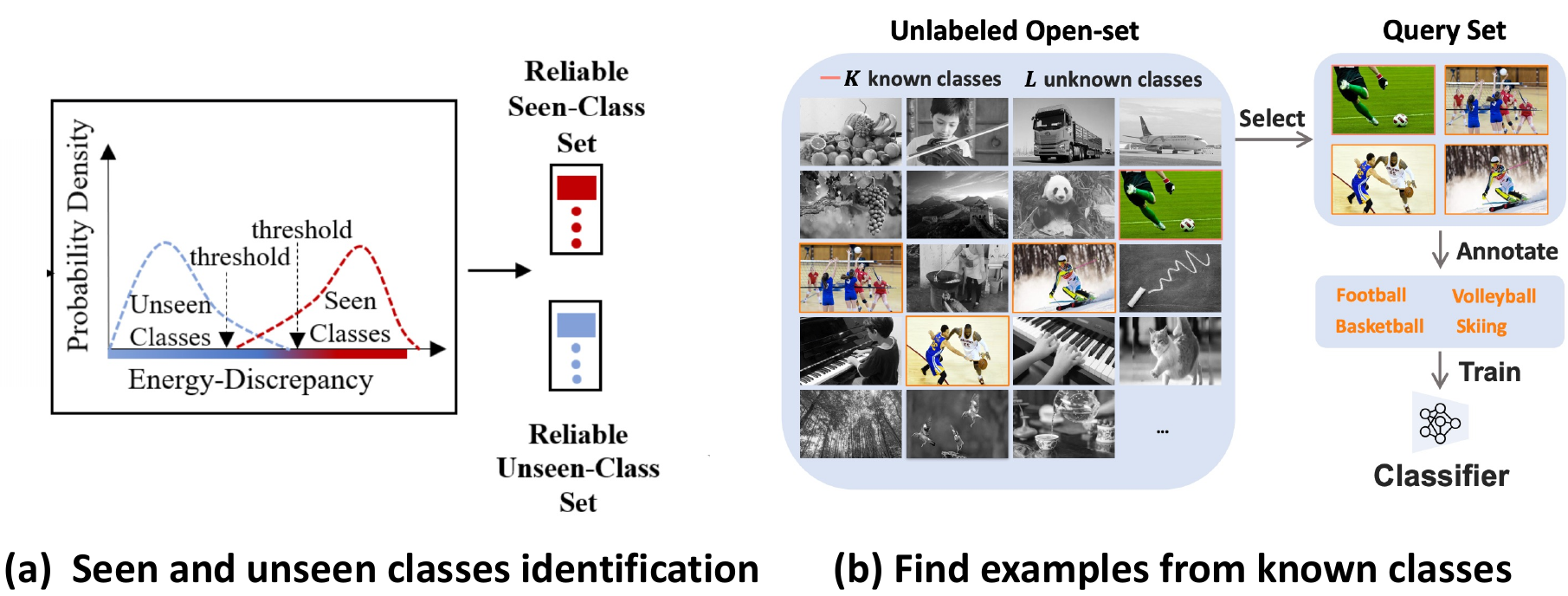}
\caption{Methods for solving class distribution mismatch.}
\label{pipeline-mismatch}
\end{figure}

To address this issue,
%
As shown in Fig.~\ref{pipeline-mismatch} (a), He et al.~\cite{DBLP:conf/cvpr/HeHLY22} propose the energy discrepancy to measure the density distribution between the seen and unseen classes. Then, they propose an iterative optimization strategy to facilitate the teacher-student distillation network to avoid selecting samples from unseen classes.
Furthermore, Tang et al.~\cite{DBLP:conf/ijcai/TangH21} propose a dual DAL framework that simultaneously performs model search and data selection. Their framework effectively addressed the issue of distribution mismatch and significantly improves model performance.
In Fig.~\ref{pipeline-mismatch} (b),  Ning et al.~\cite{DBLP:conf/cvpr/NingZLH22} introduce a detector-classifier DAL framework, where the detector filters unknown classes using Gaussian Mixture Models and the classifier selects uncertain in-distribution samples for retraining. By actively acquiring purer in-distribution query sets, this framework improves the model generalization on class distribution mismatch.

\section{Conclusion}
\label{section:conclusion}
Due to the advantages of DAL, such as high efficiency, good effectiveness, and strong robustness, DAL has been deployed in both research and industry projects.
This article provides a comprehensive survey on DAL, including
its collection, definition, influential baselines and datasets, taxonomy, applications, challenges, and some inspiring prospects.
First, we discuss the collection and filtering of DAL papers to ensure their high-quality.
Second, we give the definition of DAL tasks, and present its basic pipeline, influential baselines, and widely used datasets.
Third, we present our taxonomy for DAL methods from several perspectives and discuss their strengths and weaknesses.
From them, we obtain some guidelines for selecting different query strategies, deep model architectures, and learning paradigms to apply for different tasks. 
In addition, different annotation strategies can significantly reduce manual labor while also bringing certain drawbacks.
In terms of training process, curriculum learning training and Pre+FT can better adapt to the current era of large language models.
Fourth, we discuss some typical applications of DAL.
Other than the commonly used and popular DAL methods used for CV tasks, we also introduce the carefully designed DAL method for NLP, DM, etc.
Finally, even though DAL has many benefits, we reckon that they can be refined further in terms of pipeline, tasks, and datasets. 
Specifically, there are many problems that DAL is hard to handle, such as inefficient human annotation, difficulty in cross-domain transfer, unstable performance, lack of scalability, data imbalance, and class distribution mismatch. 
We share DAL-related resources on Github.
We hope that this work will be a quick guide for researchers and motivate them to solve important problems in the DAL domain.

\bibliographystyle{IEEEtran}
\bibliography{reference}

\begin{IEEEbiography}[{\includegraphics[width=1.1in,height=1.2in,clip,keepaspectratio]{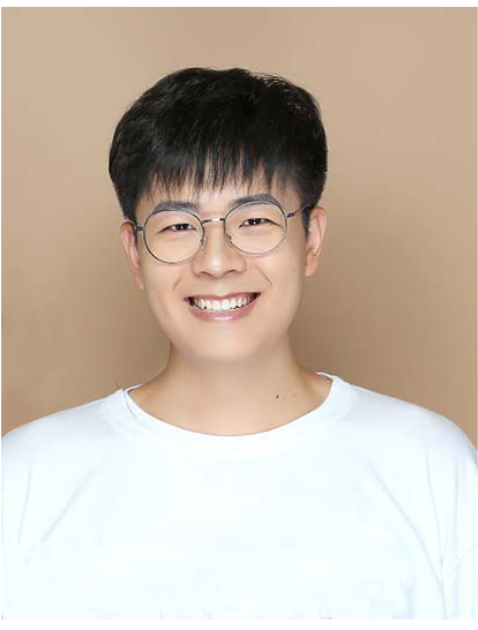}}]{Dongyuan Li} received his bachelor's degree in computer science from Dalian University of Technology in 2018. He received his Master's degree from the School of Computer Science and Technology, Xidian University in 2021.
He currently is a third-year Ph.D. candidate in the School of Information and Communication Engineering, Tokyo Institute of Technology.
His research interests include natural language processing, machine learning, data mining, social network analyses, and bio-informatics.
\end{IEEEbiography}

\begin{IEEEbiography}[{\includegraphics[width=1.1in,height=1.2in,clip,keepaspectratio]{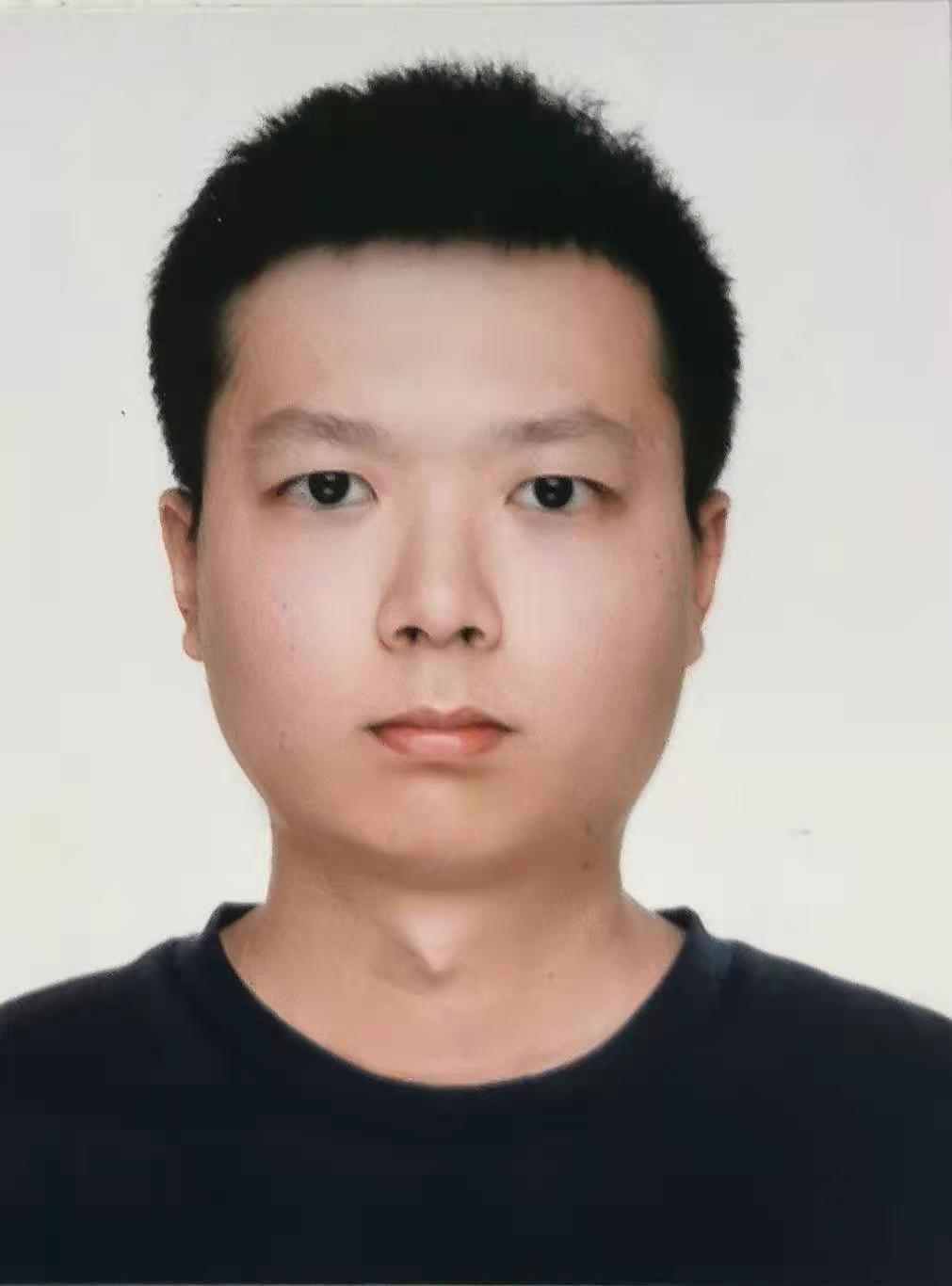}}]{Zhen Wang} received his bachelor's degree in computer science from Beihang University in 2018. He received his Master's degree from the Faculty of Electrical Engineering, Mathematics, and Computer Science, Delft University of Technology. He is currently pursuing his Ph.D. degree in the School of Information and Communication Engineering, Tokyo Institute of Technology. His research interests include natural language processing, computer vision, and multimodal learning.
\end{IEEEbiography}

\begin{IEEEbiography}[{\includegraphics[width=1.1in,height=1.2in,clip,keepaspectratio]{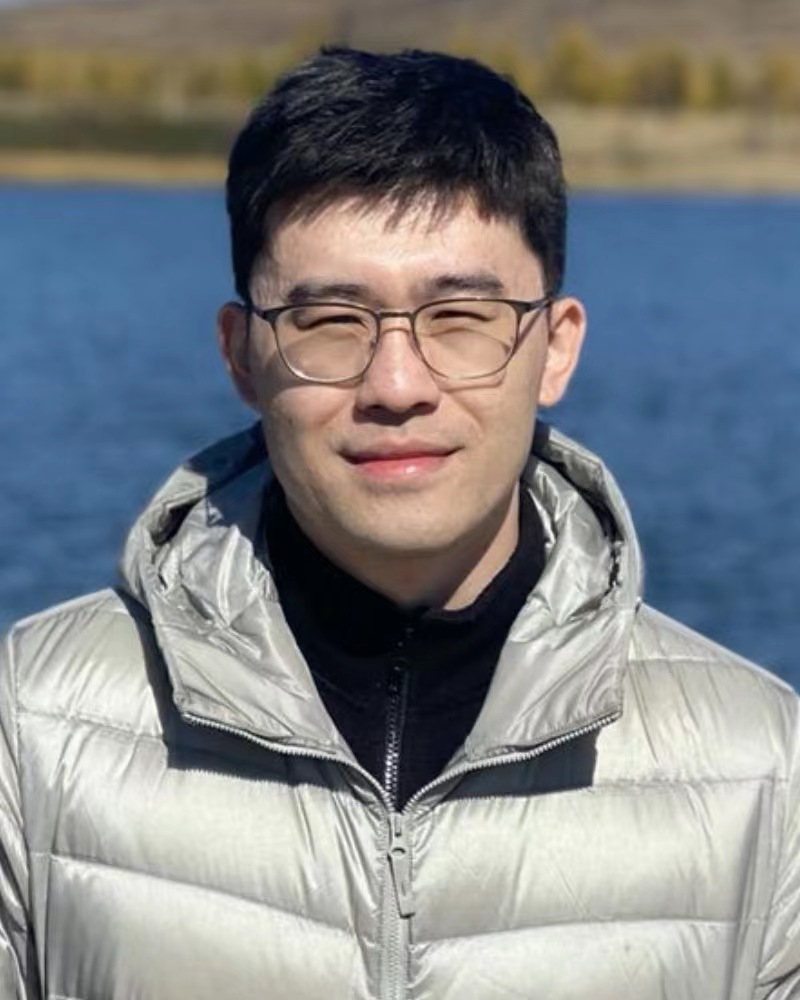}}]{Yankai Chen} currently is a fourth year Ph.D. candidate in Department of Computer Science and Engineering, The Chinese University of Hong Kong. 
He received the B.S. degree from Nanjing University in 2016 and the M.S. degree from the University of Hong Kong in 2018.
His research interests include data mining and applied machine learning and for database management and information retrieval.
\end{IEEEbiography}

\begin{IEEEbiography}[{\includegraphics[width=1.1in,height=1.2in,clip,keepaspectratio]{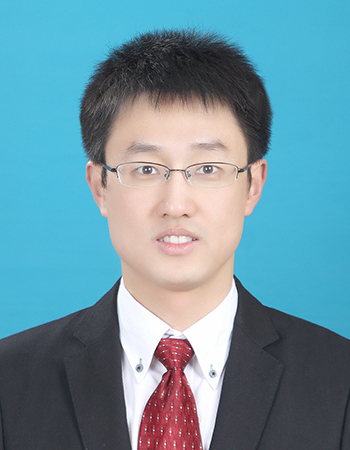}}]{Renhe Jiang}  is a lecturer at Center for Spatial Information Science, The University of Tokyo. He received his B.E. degree in Software Engineering from Dalian University of Technology in 2012, M.S. degree in Information Science from Nagoya University in 2015, and Ph.D. degree in Civil Engineering from The University of Tokyo in 2019. From 2019 to 2022, he was an assistant professor at Information Technology Center, The University of Tokyo. His research interests include spatiotemporal data mining, multivariate time series, graph neural networks.
\end{IEEEbiography}

\begin{IEEEbiography}[{\includegraphics[width=1.1in,height=1.2in,clip,keepaspectratio]{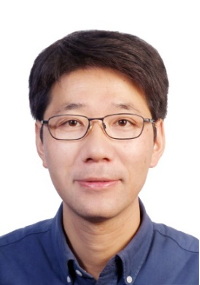}}]{Weiping Ding} (M’16-SM’19) received the Ph.D. degree in Computer Science, Nanjing University of Aeronautics and Astronautics, Nanjing, China, in 2013. In 2016, He was a Visiting Scholar at National University of Singapore, Singapore. From 2017 to 2018, he was a Visiting Professor at University of Technology Sydney, Australia. He is a Full Professor with the School of Information Science and Technology, Nantong University, Nantong, China, and also the supervisor of Ph.D postgraduate by the Faculty of Data Science at City University of Macau, China. His main research directions involve DNNs, multimodal machine learning, and medical images analysis. He has published over 250 articles, including over 110 IEEE Transactions papers. His nineteen authored/co-authored papers have been selected as ESI Highly Cited Papers. He has co-authored four books. He has holds 28 approved invention patents, including two U.S. patents and one Australian patent. He serves as an Associate Editor/Editorial Board member of IEEE Transactions on Neural Networks and Learning Systems, IEEE Transactions on Fuzzy Systems, IEEE Transactions on Intelligent Transportation Systems, IEEE Transactions on Intelligent Vehicles, IEEE Transactions on Artificial Intelligence, Information Fusion, Information Sciences, Neurocomputing, Applied Soft Computing. He is the Leading Guest Editor of Special Issues in several prestigious journals, including IEEE Transactions on Evolutionary Computation, IEEE Transactions on Fuzzy Systems, and Information Fusion. He is the Co-Editor-in-Chief of both Journal of Artificial Intelligence and Systems and Journal of Artificial Intelligence Advances.
\end{IEEEbiography}

\begin{IEEEbiography}[{\includegraphics[width=1.1in,height=1.2in,clip,keepaspectratio]{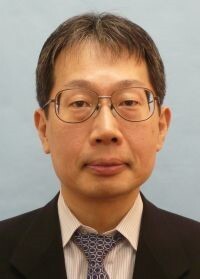}}]{Manabu Okumura} was born in 1962. He received B.E., M.E. and Dr. Eng.
from Tokyo Institute of Technology in 1984, 1986 and 1989,
respectively. He was an assistant professor at the Department of Computer Science, Tokyo Institute of Technology from 1989 to 1992, and an associate professor at the School of Information Science, Japan Advanced Institute of Science and Technology from 1992 to 2000. He was also a visiting associate professor at the Department of Computer Science, University of Toronto from 1997 to 1998. From 2000, he was an associate professor at Precision and Intelligence Laboratory, Tokyo Institute of Technology, and he is currently a professor at Institute of Innovative Research, Tokyo Institute of Technology. 
His current research interests include natural language processing, especially text summarization, computer assisted language learning, and text data mining.
\end{IEEEbiography}

\vfill

\end{document}